\documentclass[11pt, letterpaper, logo, onecolumn, numbering]{eorobotics}
\usepackage[all]{hypcap}
\usepackage[comma,authoryear,compress]{natbib}
\usepackage{hyperref}[citecolor=lightblue]

\hypersetup{
    colorlinks = true,
    citecolor = {lightblue},
    linkcolor = {lightblue},
    urlcolor = {lightblue}
}

\usepackage{algorithm}
\usepackage{algorithmicx}
\usepackage{algpseudocode}
\usepackage{microtype}
\usepackage{graphicx}
\expandafter\def\csname ver@subfig.sty\endcsname{}
\usepackage{float}
\usepackage{bigstrut}

\usepackage{amsmath}
\usepackage{amssymb}
\usepackage{mathtools}
\usepackage{amsthm}
\usepackage{mathrsfs}
\usepackage{nicefrac}
\usepackage{dsfont}
\usepackage{enumitem}
\usepackage{subcaption}
\usepackage{graphicx,subfig}
\usepackage{float}
\usepackage[utf8]{inputenc} %
\usepackage[T1]{fontenc}    %
\usepackage{hyperref}       %
\usepackage{url}            %
\usepackage{booktabs}       %
\usepackage{amsfonts}       %
\usepackage{nicefrac}       %
\usepackage{microtype}      %
\usepackage{fdsymbol}
\usepackage{wrapfig}
\usepackage{lipsum}
\usepackage{stackengine}
\usepackage[font=small,labelfont=bf]{caption}
\usepackage{color}
\usepackage{adjustbox}
\usepackage{rotating}
\usepackage{makecell}

\usepackage{xspace}
\usepackage{tikz}
\usepackage{multirow}    
\usepackage{array}       
\usepackage{colortbl}
\usepackage{tabularx}

\usepackage{pifont}   

\newtcolorbox{AIbox}[2][]{aibox,title=#2,#1}
\definecolor{lightblue}{rgb}{0.22,0.45,0.70}%
\definecolor{Gray}{gray}{0.95}
\definecolor{Cornsilk}{rgb}{1.0, 0.97, 0.86}

\usepackage{amsmath}

\usepackage[all]{hypcap}

\usepackage[capitalize,noabbrev]{cleveref}
\crefname{section}{Section}{\S\S}
\Crefname{section}{Section}{\S\S}
\crefname{table}{Table}{Tables}
\crefname{figure}{Figure}{Figures}
\crefname{algorithm}{Algorithm}{}
\crefname{equation}{eq.}{}
\crefname{appendix}{Appendix}{}
\crefformat{section}{Section #2#1#3}

\usepackage{multicol}

\usepackage{titlesec}

\newcommand{\ours}{\textbf{\scalebox{1.0}{EO}-1}\xspace}
\newcommand{\dataset}{\textbf{\scalebox{1.0}{EO}-Data1.5M}\xspace}
\newcommand{\benchmark}{\textbf{\scalebox{1.0}{EO}-Bench}\xspace}

\definecolor{deltaBg}{RGB}{220,230,255} 

\newcommand{\rowhighlight}{\rowcolor{deltaBg}}

\def \H1{H1}
\def \G1{G1}

\title{\bf EO-1: An Open Unified Embodied Foundation Model for General Robot Control}

\correspondingauthor{
    Shanghai AI Laboratory;
    Fudan University;
    AgiBot;
    Northwestern Polytechnical University \\
    $\dagger$ Project Leaders, Email \href{mailto:wangdong@pjlab.org.cn}{wangdong@pjlab.org.cn}
}

\author{Delin Qu$^\ast$, Haoming Song$^\ast$, Qizhi Chen$^\ast$, Zhaoqing Chen$^\ast$, Xianqiang Gao$^\ast$, Dong Wang$\dagger$, Modi Shi, Guanghui Ren Maoqing Yao, Bin Zhao$\dagger$, Xuelong Li}

\begin{document}
\begin{abstract}
The human ability to seamlessly perform multimodal reasoning and physical interaction in the open world is a core goal for general-purpose embodied intelligent systems. Recent vision-language-action (VLA) models, which are co-trained on large-scale robot and visual-text data, have demonstrated notable progress in general robot control. However, they still fail to achieve human-level flexibility in interleaved reasoning and interaction. 
In this work, we introduce \textbf{EO-Robotics}, consists of \ours model and \dataset dataset. \ours is a unified embodied foundation model that achieves superior performance in multimodal embodied reasoning and robot control through interleaved vision-text-action pre-training. The development of \ours is based on two key pillars: (i) a unified architecture that processes multimodal inputs indiscriminately (image, text, video, and action), and (ii) a massive, high-quality multimodal embodied reasoning dataset, \dataset, which contains over 1.5 million samples with emphasis on interleaved vision-text-action comprehension. \ours is trained through synergies between auto-regressive decoding and flow matching denoising on \dataset, enabling seamless robot action generation and multimodal embodied reasoning. Extensive experiments demonstrate the effectiveness of interleaved vision-text-action learning for open-world understanding and generalization, validated through a variety of long-horizon, dexterous manipulation tasks across multiple embodiments. This paper details the architecture of \ours, the data construction strategy of \dataset, and the training methodology, offering valuable insights for developing advanced embodied foundation models.
\end{abstract}
\maketitle

\section{Introduction}
\label{sec:intro}
Generalist robot policies with open-world capabilities is essential for deploying autonomous robots in real scenarios, where they need to tackle diverse tasks, follow various instructions, adapt to different situations, and even handle unforeseen events. In the realm of digit domains, the paradigm of learning from large-scale multimodal datasets has demonstrated broad proficiency and generalization across various interactive and assistive applications. Extending this idea to the physical world, researchers are now training large vision-language-action models on extensive robotic datasets to realize generalist robots with similar versatility. Although recent efforts have demonstrated impressive progress in achieving dexterous manipulation and long-horizon physical control within constrained environments, open world generalization remains the central challenge for developing general-purpose autonomous AI agents in the physical world~\citep{intelligence2025pi_0_5,gemini_robotics}. To address this challenge, robots must be capable of acquiring comprehensive world knowledge, engaging in human-level reasoning, and executing dexterous actions.


Early generalist robot policies~\citep{kim2024openvla,black2024pi_0,pertsch2025fast} primarily extend vision–language models (VLMs) into vision-language-action (VLA) models with domain-specific robotic data, either through auto-regressive decoding of discrete action tokens or by incorporating additional continuous flow matching modules. However, because these VLA models are trained exclusively on robotic datasets, they are restricted to narrow task domains and specific environments. As a result, they suffer from diminished general semantic knowledge inherited from VLMs and exhibit limited instruction-following capabilities. 

Recently, several studies~\citep{intelligence2025pi_0_5,driess2025knowledge,lin2025onetwovla} have explored co-training VLA models with both web data and robotic data, showing promising generality when interacting with new objects and unseen backgrounds.
Nevertheless, existing approaches remain predominantly generating robot action at the end of the VLA output sequence, overlooking the rich temporal dynamics and causal dependencies among vision, language, and action modalities inherent in open-world embodied interactions.
Humans, in contrast, exhibit a flexible and interleaved synergy between multimodal embodied reasoning and physical action, enabling highly generalizable and dexterous manipulation in the open world, \emph{i.e.}, reasoning guides action and action results inform subsequent reasoning.
This raises a fundamental research question: \textbf{How can we design an effective training paradigm for generalist robot policies that support flexible and mutual-informed reasoning-acting integration?}

Inspired by the impressive results of advanced multimodal understanding and generation systems~\citep{deng2025emerging,ma2025janusflow,xie2025show,team2025nextstep}, which demonstrate the superiority in interleaved multimodal pretraining, we propose a unified embodied model to enable flexible and powerful multimodal embodied reasoning and action generation through interleaved embodied pretraining. Achieving this requires not only an effective and unified architecture that excels in mix-modality generation, but also carefully structured multimodal embodied data that jointly integrate texts, images, videos, and robotic actions. 

To realize this vision, we first established a new protocol for scalable data curating, filtering, and construction of high-quality multimodal interleaved embodied data. As for data source, we integrate web vision-language data with real robot episodes, the latter naturally providing action-level, temporal, and physical continuity. Then, we employ VLMs and human to annotate real robot episodes with diverse embodied temporal and spatial QA pairs, including physical common sense understanding, task planning, object localization, affordance pointing, and multi-view correspondence. These annotations enable the learning of fine-grained geometric and spatial-temporal representations of the physical world. Finally, the interleaved vision-text-action data is constructed by concatenating these multimodal QA and robot control actions in temporal order. Importantly, while robot actions are fixed at each timestep, we design three flexible interleaved formats to associated random embodied reasoning QA pairs. 
Consequently, the interleaved embodied data capture rich world knowledge and nuanced cross-modal interactions, providing models with foundational capabilities of action prediction, scene understanding, and complex multimodal reasoning for generalist robot policies.

Regarding model architecture, we adopt a single unified decoder-only transformer that integrates discrete auto-regressive decoding with continuous flow matching denoising. The unified model is built on a pre-trained VLM, thereby inheriting broad visual–language knowledge, and its shared parameters are further optimized with modality-specific objectives: next-token prediction for text and flow matching for robotic actions.
The model applies causal attention cross the entire interleaved vision-text-action sequence to capture the sequential dependencies between reasoning and acting. In addition, two MLPs are added to encode and decode continuous robotic actions, complementing the original text and visual tokenizers.
Unlike prior VLA models~\citep{black2024pi_0,intelligence2025pi_0_5,driess2025knowledge} that introduce extra action-specific modules to learn action generation, our design enables easier alignment between vision/language and action modalities through circumventing train new action-specific parameters from scratch, thereby achieving more effective cross-modal knowledge transfer in generalist robot policies.

Overall, we present the EO-Robotics toolchain for advanced embodied foundation model development, where \ours, a unified embodied foundation model comprising 3B parameters, is trained on the carefully curated interleaved embodied dataset \dataset. As a generalist VLA, \ours exhibits strong generalization capabilities in multimodal embodied reasoning and real robot control across a diverse set of challenging tasks, including pick-and-place, articulated manipulation, long-horizon planning and dexterous skills. The impressive results validate the effectiveness of unified multimodal modeling for generalist embodied intelligence.
Furthermore, EO-Robotics is released with full openness, including model weights, training code, and all components of the interleaved embodied dataset, to serve as a forward step for general-purpose automatous robots and facilitate further research in the community. This paper highlights the following contributions:
\begin{itemize}[leftmargin=*, noitemsep]
    \item \textbf{Unified Architecture:}
          \ours is a unified model that integrates multimodal embodied reasoning and real robot control within a shared backbone, enabling seamless cross-modal interaction without introducing extra bottleneck components or action-specific parameters.

    \item \textbf{Interleaved Embodied Dataset:} 
     \dataset is a comprehensive dataset derived from the largest robotic datasets, featuring interleaved embodied reasoning and and robot control through a scalable data construction pipeline, thereby enabling interleaved vision-text-action pretraining. 
    \item \textbf{Real-world Generalization:} Our model surpasses existing open-source models across multiple embodied reasoning and robot control benchmarks, including ERQA, LIBERO, SimplerEnv, and the self-constructed 
 \benchmark, meanwhile, extensive real-robot evaluations demonstrate its substantially stronger reasoning capabilities and dexterous control in open-world generalization. 
\end{itemize}
\section{The \ours Model and Training Paradigm}
\label{sec:method}
\begin{figure*}[t]
    \centering
    \includegraphics[width= 1\linewidth]{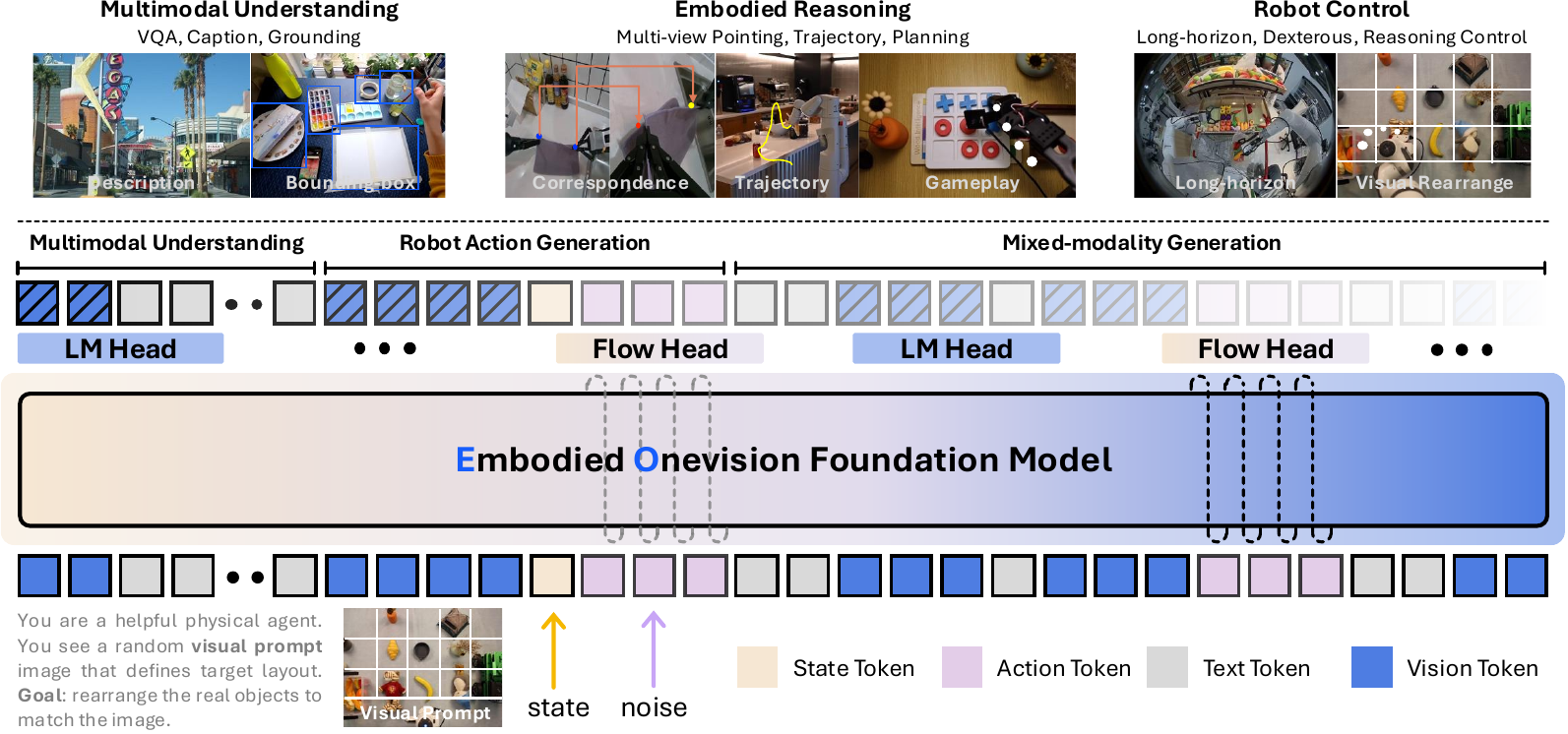}
    \caption{\textbf{\ours Model Architecture}. \ours model is a Vision-Language-Action (VLA) model that adopts a single unified decoder-only transformer, equipping with discrete language-modeling head for multimodal embodied reasoning and continuous flow-matching head for robot action generation. The language instruction, image observations, robot state, and noisy action are encoded into an interleaved token sequence of tokens to be processed by the shared transformer backbone, whose weights are initialized from Qwen2.5-VL. The model is trained on interleaved vision-text-action data with a combination of flow-matching objective and next-token-prediction objective and capable of seamless embodied reasoning and acting.}
    \label{fig:pipeline}
\end{figure*}

In this section, we introduce the architecture of \ours and its training strategies. As illustrated in Figure~\ref{fig:pipeline}, the main architectural idea of \ours is to capture the inherent temporal dynamics and causal relationships between vision-text-action modalities in embodied interactions.
This is realized by modeling multimodal understanding and robot control with a single unified decoder-only transformer, enabling effective cross-modal knowledge transfer and alignment for generalist policies.

\subsection{Problem Formulation}

We formulate the generalist robot policy $\pi_\theta$ as a unified model \ours that integrates \textit{Multimodal Embodied Reasoning} and \textit{Robot Control} through a synergized autoregressive and denoising paradigm. 
The \ours architecture can flexibly decode both continuous action chunks and tokenized text outputs, enabling seamless text-based reasoning and physical robot control within one model.
The distribution represented by the model can be written as $ \pi_\theta(\hat{\mathbf{a}}_{t:t+h}, \hat{\mathbf{\ell}_t}  \vert {\mathbf{o}_t, \mathbf{\ell}_t, \mathbf{a}_{t-h:t}})$, where $\mathbf{o}_t = [\mathbf{I}_t^1,...,\mathbf{I}_t^n, \mathbf{q}_t]$ consists of multi-view image observations and robot state, language contexts $\ell$ are embodied reasoning question-answer text data with respect to current observation (\emph{e.g.}, ``Q: Based on current image observation, what is the next step to do for the robot cleaning the table. A: pick up the yellow box and place it in the trash box'') or overall task prompts (\emph{e.g.}, ``clear the table''), and action sequence ${\mathbf{a}}_{t:t+h}$ is an action chunk. 
The unified model is trained on an interleaved multimodal dataset $\mathcal{D}$ to alternately generate text $\hat{\mathbf{\ell}}_t$ and robot action $\hat{\mathbf{a}}_{t:t+h}$ for seamless embodied reasoning and action. Note that the unified model does not generate outputs on vision tokens and robot state tokens.

The generalist robot policy $\pi_\theta$ is instantiated by a transformer that takes interleaved multimodal token sequence $x_{1:N}$ as input and generates a sequence of multimodal outputs $\hat{x}_{1:N}$ in an autoregressive manner, \emph{i.e.,}  ${p}(\hat{x}_{1:N})=\prod_{i=1}^N{p}(\hat{x}_{i} | x_{<i})$. For inputs, each token can be a discretized text token $x_i^w$, a continuous image patch token $x_i^I$, a continuous robot state token $x_i^q$, or a partially denoised robot action token $x_i^a$. For output, discrete text tokens are sampled via a language-modeling head, while continuous action tokens are sampled by a flow-matching head.
The language modeling head is implemented with a classic logits head appended to the unified transformer to decode text token output $\hat{x}_i^w$. 
For the flow-matching head, rectified flow~\citep{lipman2022flow} is utilized to generate continuous action signals $\hat{\mathbf{a}}_t$ according to the forward Euler integration rule:
\begin{equation}
    \hat{\mathbf{a}}_t^{\tau+\delta} = \hat{\mathbf{a}}_t^{\tau} + \delta V_{\pi_{\theta}}(\hat{\mathbf{a}}_t^{\tau}, \vert {\mathbf{o}_t, \mathbf{\ell}_t, \mathbf{a}_{t-h:t}}),
\end{equation}
where $V_{\pi_{\theta}}$ represents the parameters of $\pi_{\theta}$ to predict the vector field for robot action generation. The actions are generated by integrating the predicted vector field from $\tau=0$ to $\tau=1$, starting with random noise ${z}_t^0 \sim \mathcal{N}(\mathbf{0}, \mathbf{I})$, and $\delta$ is the integration step size. The shared parameters in the unified transformer enable the seamless transfer of semantic knowledge from vision-language understanding to action generation, leading to more general reasoning capabilities and control capabilities in embodied scenarios~\citep{deng2025emerging,intelligence2025pi_0_5,driess2025knowledge}.

\subsection{Model and Training}
\paragraph{Model Architecture.} The proposed unified model architecture processes interleaved multimodal inputs (\emph{i.e.},  text, images, videos, and actions) through a shared transformer backbone that generates both discrete token sequences and continuous action tokens. The model employs a text tokenizer and visual encoder to convert text and image patches into input tokens, while the robot state $\mathbf{q}_t$ is linearly projected into the same transformer embedding space, \emph{i.e.,} $x_i^{w}, x_i^{I}, x_i^{q} \in \mathbb{R}^{d}$. Note that the text tokenizer and visual encoder are inherited from pre-trained VLM and the state projector is random initialized. For flow matching action denoising, the input ``noisy action'' $\mathbf{a}_t^\tau$ is obtained by $\mathbf{a}_t^\tau=\tau\mathbf{a}_t+(1-\tau)z_\tau$, where $z_\tau \sim \mathcal{N}(0, I)$ is random noise. Then, another noisy action linear projector is utilized to embed noisy action and flow matching timestep $\tau$ into noisy action token $x_i^a \in \mathbb{R}^{d}$. The unified transformer backbone, initialized from Qwen 2.5 VL~\citep{qwen2_5_vl}, processes this interleaved multimodal input sequence with shared parameters and generates output sequence through two separate heads: a language head for text token decoding and a flow head for continuous action denoising. Note that we do not introduce extra action-specific parameters to separately model action denoising as prior VLA models, enabling seamless integration of discrete multimodal embodied reasoning and continuous robot control. 

\begin{figure}[t]
    \centering
    \includegraphics[width=1.0\linewidth]{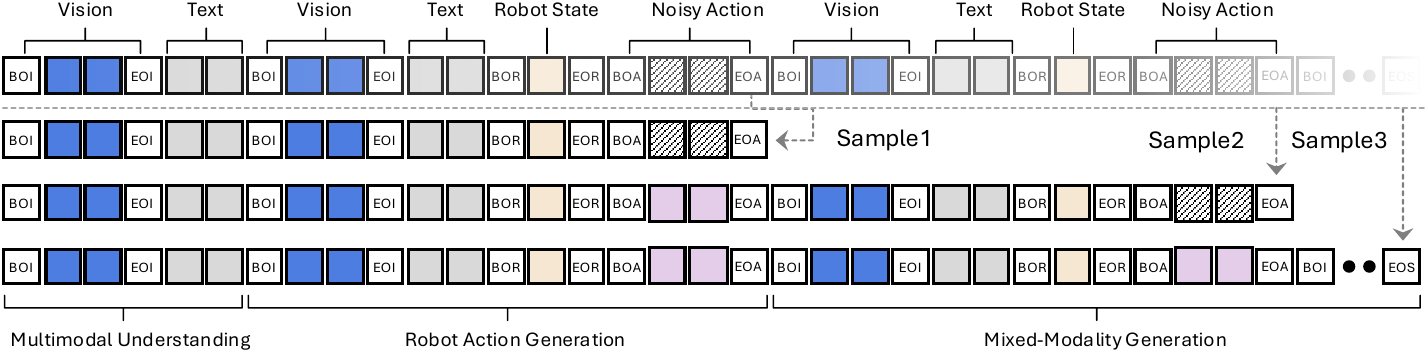}
    \caption{\textbf{Interleaved rectifying sampling strategy.} Our method samples variable-length subsequences from robot action generation segments, enabling efficient training of mixed-modality generation while preserving causal relationships.}
    \label{fig:interleaved_rectifying_sampling}
    \vspace{-2ex}
\end{figure}
\noindent\textbf{Unified Prompting and Attention.} 
Training \ours~involves three types of data: multimodal understanding data, robot action generation data, and mixed-modality generation data (\cref{sec:data}). The mixed-modality generation data is formatted as interleaved vision-text-action data, \emph{i.e.}, ``\texttt{[BOI]} \{image patchs\} \texttt{[EOI]} \texttt{[BOS]} \{text\} {\texttt{[EOS]} \texttt{[BOR]} \{robot state\} \texttt{[EOR]} \texttt{[BOA]} \{noisy action\} \texttt{[EOA]} \texttt{[BOI]} \{image patchs\} \texttt{[EOI]} \texttt{[BOS]} \{text\} {\texttt{[EOS]} \texttt{[BOR]} $\cdots$'', where \texttt{[BOS]}, \texttt{[EOS]}, \texttt{[BOI]}, \texttt{[EOI]}, \texttt{[BOR]}, \texttt{[EOR]}, \texttt{[BOA]}, and \texttt{[EOA]} denote the beginning and end of sentence, image, robot state, action, respectively. The multimodal understanding format is ``\texttt{[BOI]} \{image patchs\} \texttt{[EOI]} \texttt{[BOS]} \{text\} {\texttt{[EOS]}'' and the robot control data format is ``\texttt{[BOI]} \{image patchs\} \texttt{[EOI]} \texttt{[BOS]} \{text\} {\texttt{[EOS]} \texttt{[BOR]} \{robot state\} \texttt{[EOR]} \texttt{[BOA]} \{noisy action\} \texttt{[EOA]}''. During training, we employ an omnidirectional attention mask mechanism to process these three type data with causal attention masks and cache key-value (KV) pairs of the generated multimodal context to accelerate mixed-modality generation at inference. 
Note that prior VLA models are either trained with solo robot control data or co-trained with multimodal understanding and robot control data, missing interleaved vision-text-action data.


\noindent\textbf{Interleaved Rectifying Sampling.} 
When training on interleaved vision-text-action data for mixed-modality generation, the denoising process of action generation in interleaved data can disrupt the causal relationships in multimodal token sequences, because the subsequent text, image, or action tokens should attend to the clean action tokens and preceding text/image tokens, rather than noisy action tokens. To address this challenge, we propose a rectifying sampling strategy for mixed-modality generation training.
As illustrated in Figure~\ref{fig:interleaved_rectifying_sampling}, we sample $N+1$ (\emph{e.g.}, $N=2$) training subsequences from an interleaved sequence containing $N$ action generation segments by splitting the sequence with respect to continuous action generation segments.
For the subsequences ending with the first action generation segment, there is no additional operation on it.
For the subsequences ending with the interleaved action generation segment, \emph{i.e.}, containing another action generation segment in the middle of the sequence, we replace the noisy action tokens in the middle action generation segment with clean action tokens.
With this interleaved rectifying sampling, each action generation segment is trained with both flow-matching denoising on noisy action tokens and indirect gradients backpropagation on clear action tokens for subsequent interleaved generation.
%


\noindent\textbf{Training Objective.}
To learn both auto-regressive decoding and flow-matching denoising with one unified model, we employ two learning objectives: i) next token prediction and ii) denoising vector field prediction.
For multimodal understanding and embodied reasoning tasks, the output token $\hat{x}_i$ contains only text tokens $\hat{x}^w_i$ and is obtained through original language decoding head. The language head and transformer backbone are trained on decoded text tokens using the cross-entropy loss between ground-truth text tokens and predicted logits $\mathcal{L}_{ar}(x_{gt}^w, \hat{x}^w_i)$, where $\hat{x}^w_i=\pi_{\theta}(\hat{x} |x_{<i})$ is conditioning on all preceding multimodal tokens.
For robot action generation, the output action $\hat{\mathbf{a}}_t$ is generated by mapping intermediate hidden features of the last transformer layer to denoise vector field $z_0-\mathbf{a}_t$ through a separate flow head. The flow head and transformer backbone are trained on action tokens by minimizing the following loss:
\begin{equation}
    \mathcal{L}_{fm}(\theta)=\mathbb{E}_\tau[\left\|V_\theta\left(\mathbf{a}_t^\tau, \tau | x_{<a})-\left(z_0-\mathbf{a}_t\right)\right\|^2\right],
\end{equation}
where $\mathbf{a}_t^\tau = \tau \mathbf{a}_t + (1-\tau) z_0$ is the input noisy action. $x_{<a}$ represents the all preceding multimodal tokens.
As in~\citep{black2024pi_0}, we sample the flow matching timestep $\tau$ from a beta distribution that emphasizes lower (noisier) timesteps. With the interleaved multimodal token sequence, we  apply next token prediction $\mathcal{L}_{ar}(\theta)$ to the language head and flow matching $\mathcal{L}_{fm}(\theta)$ to the flow head for predicting denoising vector field. The unified model is trained end-to-end by optimizing sum of these two objectives: $\mathcal{L} = \mathcal{L}_{ar}(\theta) + \mathcal{L}_{fm}(\theta)$.
\section{Dataset and Benchmark}
\label{sec:data}
\ours is trained on a diverse range of datasets across multiple modalities, including text, image, video, and robot control data, to perform embodied reasoning and dextrous control, all through a unified multimodal interface. In addition to standard robot control datasets and existing large-scale vision-language datasets, we design a scalable data construction pipeline and build interleaved embodied datasets from large-scale robot control episodes to capture the rich temporal dynamics and causal relationships inherent in embodied interactions. As illustrated in~\cref{tab:data_overview}, The pre-training data corpus is structured into three main categories: web multimodal data, robot control data, and interleaved embodied data. We summarize the scale and composition of our training data across different modalities. In the following sections, we detail our data sources, interleaved data construction pipeline, and show some samples used in training.
\begin{table*}[!h]
    \centering
    \resizebox{1.0\textwidth}{!}{
        \begin{tabular}{llcr}
            \toprule
            \bf Modality              & \bf Source                                                                                                                                                                                                                                    & \bf \#Data     & \bf \#Tokens \\
            \midrule
            Web Multimodal Data       & \begin{tabular}[c]{@{}l@{}}LLaVA-1.5~\citep{liu2024improved}, LLaVA-Video-178K~\citep{zhang2024video}, \\PixMo-Points~\citep{molmo2024}, RefCOCO~\citep{kazemzadeh-etal-2014-referitgame}, \\RoboVQA~\citep{sermanet2024robovqa}\end{tabular} & 5.7M           & 7.1B         \\
            Robot Control Data        & \begin{tabular}[c]{@{}l@{}}AgiBotWorld~\citep{bu2025agibot}, Open X-Embodiment~\citep{o2024open}, \\RoboMIND~\citep{wu2024robomind}, SO100-Community~\citep{shukor2025smolvla},\\IPEC-Franka~\citep{qu2025spatialvla}\end{tabular}             & 1.2M (episode) & 127.3B       \\
            \rowhighlight
            Interleaved Embodied Data & \dataset from AgiBotWorld, Open X-Embodiment and RoboMIND                                                                                                                                                                                      & 1.5M           & 1.0B         \\
            \bottomrule
        \end{tabular}
    }
    \caption{Overview of the data used in \ours training. More details in \cref{sec:training_dataset_statistics}.}
    \label{tab:data_overview}
\end{table*}

\subsection{Data Source}
\noindent{\textbf{Web Multimodal Data.}}
The web text-image paired data is essential for multimodal understanding. We curated a web multimodal dataset comprising \textit{5.7 Million samples with 7.1 Billion tokens}, integrating six public sources across three task domains: i)
\textbf{Visual instruction following.} We incorporate the \textit{LLaVA Visual Instruct Series}~\citep{liu2023llava,sermanet2024robovqa} datasets, which contain GPT-generated multimodal instructions and robotics-oriented visual question answering examples. These sources cover diverse interaction patterns, including image-grounded conversation, multi-step reasoning, and long-horizon task planning.
ii) \textbf{Video understanding and reasoning.} We leverage the \textit{LLaVA-Video}~\citep{zhang2024llavanext-video} and \textit{RoboVQA}~\citep{sermanet2024robovqa}, which support temporal multimodal comprehension through detailed video captioning, open-ended QA, and multiple-choice question answering.
iii) \textbf{Referring expression and spatial grounding.} To support fine-grained object understanding, we include \textit{RefCOCO (subset)}~\citep{refcoco} for natural language referring expression comprehension and \textit{PixMo-Points}~\citep{pixmo} for spatial localization via human-annotated coordinate grounding.

\noindent{\textbf{Robot Control Data.}}
Robot control data provides critical supervision for policy learning across embodiment generalization and fine-grained manipulation. We aggregate a large-scale real robot control dataset comprising \textit{1.2 Million episodes with 0.13 Trillion tokens} from five public sources, forming one of the most comprehensive collections for robot learning to date. The dataset includes \textit{AgiBot-World}~\citep{bu2025agibot}, featuring dual-arm humanoid demonstrations across diverse bimanual manipulation tasks with rich visuomotor sequences; \textit{Open X-Embodiment}~\citep{o2024open}, which spans 22 robot types and 527 skills to support cross-platform generalization; and \textit{SO100-Community}~\citep{shukor2025smolvla}, a collection of long-horizon community demonstrations using SO100 platform. It also incorporates \textit{RoboMind}~\citep{wu2024robomind}, covering a wide range of tasks across multiple embodiments (e.g., Panda, UR5e, AgileX, humanoids) with diverse object classes. 
\subsection{Interleaved Embodied Data Construction}
\label{sec:data_construction}
Our interleaved embodied data is obtained by annotating existing real robot control data on two aspects, resulting in two components: i) embodied temporal and spatial reasoning data focusing on physical dynamic and spatial relationship understanding of robot execution video (as shown in~\cref{fig:statistic}) and ii) interleaved vision-text-action data connecting temporal/spatial reasoning data with robot control data for learning multimodal causal relationships in embodied interactions.
\subsubsection{Embodied Reasoning Data Pipeline}
To enhance the model's vision-language capabilities on embodied intelligence, we develop a specialized pipeline in~\cref{fig:statistic}, to carefully curate robot-specific question-answer data for both temporal and spatial embodied reasoning. Real robot manipulation videos serve as our primary data source. They perform a variety of tasks in real-world environments and showcase multiple embodiments. To ensure our curated dataset is rich in terms of diversity, quality, embodiments, and scenarios, the curation pipeline are presented as follows:
\begin{enumerate}[leftmargin=*, topsep=0pt, noitemsep]
    \item \textbf{Robot Video Filter and Curation.} As illustrated in~\cref{fig:statistic} and~\cref{sec:data_preprocessing}, we first filter and sample a set of robot videos based on visual similarity to improve the data source diversity. This is because existing robot control datasets comprise thousands of videos collected in several fixed environments performing limited tasks, resulting in high visual similarity between videos. We extract visual features with a pretrained vision backbone and cluster them according to feature similarity. The diverse video set is curated by sampling a fixed number of videos from each cluster.
    \item \textbf{Video Splitting and Captioning.} We employ both human annotators and pretrained VLMs to split videos into short clips containing individual subtasks. Each clip is then processed to extract detailed descriptions of the robot's actions. These captions are not only used to construct video captioning QA data, but also serve as prompts for subsequent embodied reasoning annotations.
    \item \textbf{Generating QA Pairs.} We prompt VLM to construct free-form questions according to designed templates, using subtask clips with previously generated captions. We construct two kinds of questions: i) ``temporal reasoning'' questions that focus on task planning and physical common sense understanding, and ii) ``spatial reasoning'' questions that require spatial understanding and object referring. As for answers, we first adopt VLM to generate multiple answers and employ human annotators to select or rewrite the correct answers.
    \begin{figure}[H]     
        \hspace*{-0.045\textwidth}
        \includegraphics[width=1.08\textwidth]{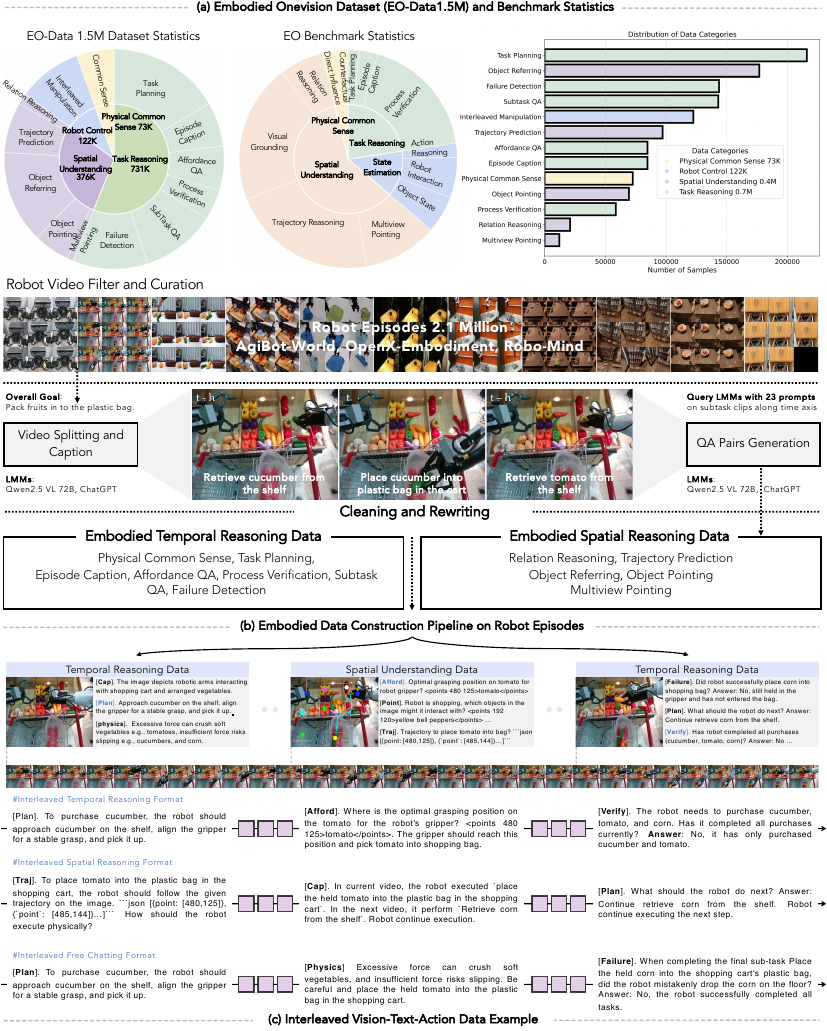}
        \caption{(a) Statistics of EO-Robotics Dataset (\dataset) and Benchmark (\benchmark). (b) Dataset curation pipeline on robot episodes. (c) Interleaved Vision-Text-Action Data Example, including three interleaved formats to concatenate embodied reasoning QA pairs and robot control data along temporal order. }
        \label{fig:statistic}
    \end{figure}
    \item \textbf{Cleaning and Rewriting.} Finally, we apply rule-based cleaning to ensure valid answers in spatial reasoning, and prompt an LLM to rewrite question-answer pairs to improve text diversity and inject fuzzy semantics into data.
\end{enumerate}

Using the aforementioned pipeline, we curated embodied temporal and spatial reasoning datasets specializing on temporal dynamic understanding and spatial relationship reasoning. The annotation details and dataset statistics are presented as below:
\noindent\textbf{Embodied Temporal Reasoning Data.}
The temporal reasoning data aims to equip the pre-trained vision-language model with capabilities on physical common sense understanding and task planning, specializing on physical commonsense reasoning and task planning.
As shown in~\cref{fig:statistic} (a), the \textbf{Physical Commonsense Understanding} data contain 70 thousand QA pairs for describing the effect of various robot actions in the physical world.
The task reasoning data comprises 0.7 million QA pairs across six sub-categories to develop the model's capacities including i) \textbf{Task Planning}: generating subtask sequences to complete long-horizon tasks. ii) \textbf{Episode Captioning}:  captioning of robot video clips. iii) \textbf{Affordance QA}: assess whether executing a specific action is possible. iv) \textbf{Process Verification}: recognizing completed actions in video clips. v) \textbf{Subtask QA}: determining whether a subtask has been successfully completed. vi) \textbf{Failure Detection}: identifying unsuccessful subtask executions.~\cref{sec:temporal_prompts} shows question-construction prompt templates used to produce these questions and answers. 

\noindent\textbf{Embodied Spatial Reasoning Data}
This kind of data focus on enhancing the model's abilities in spatial understanding and reasoning, which contains 1.5 million QA pairs and is organized into five sub-categories: relation reasoning, trajectory prediction, object referring, object pointing, and multiview pointing, as illustrated in~\cref{fig:statistic} (a). Specifically, i) \textbf{Relation Reasoning} data are annotated to understand relative spatial relationships among multiple objects in the scene. ii) \textbf{Trajectory Prediction} focuses on anticipating the future motion trajectories of objects or robot gripper within dynamic environments. iii) \textbf{Object Referring} data are constructed by grounding referred objects with bounding boxes among multiple candidates. iv) \textbf{Object Pointing} data aim to identify and point specific objects in multi-object scenarios according to the task instruction. v) \textbf{Multiview Pointing} is labeled by locating the same objects across different view images of the same scene. The question-construction prompt for embodied spatial reasoning is shown in~\cref{sec:spatial_prompts}. 

\subsubsection{Interleaved Vision-Text-Action Data Pipeline}
To learn the natural sequential coherence among visual, text, and action modalities, we design three flexible formats to concatenate embodied reasoning QA data and robot control data along temporal order in robot videos. They are illustrated in~\cref{fig:statistic} (b), and detailed as follows:
\begin{itemize}[leftmargin=*, topsep=0pt, noitemsep]
    \item \textbf{Interleaved Temporal Reasoning Format.} For a certain frame from the robot video, the interleaved temporal reasoning data is constructed as.
    ``$\cdots$ [image tokens] [next subtask plannning QA] [subtask instruction] [robot action] [image tokens] [task-completion verification QA] $\cdots$''.
    We use a predefined template merge next subtask plannning answer into the instruction for  action generation, and a task-completion verification QA is appended to new image tokens to determine the task progress.
    \item \textbf{Interleaved Spatial Reasoning Format.} As for spatial reasoning data, we use the trajectory prediction QA and introduce a ``[trajectory instruction]'' to connect spatial reasoning QA with robot control data, \emph{i.e.}, 
    ``$\cdots$ [image tokens] [robot trajectory prediction QA] [trajectory instruction] [robot action] [image tokens] [task-completion verification QA] $\cdots$''.
    The robot trajectory prediction's answers contain a sequence of [x,y] coordinates of robot gripper trajectory for finish task, and 
    its results are merged into ``[trajectory instruction]'' for next action generation, \emph{i.e.}, ``How should the robot execute the trajectory [x1,y1],$\cdots$,[x6,y6] physically?'. 
    \item \textbf{Interleaved Free Chatting Format.} For other temporal and spatial reasoning QA data, we randomly select QA pairs and connect them with robot control data through original task instruction: 
    ``$\cdots$ [image tokens] [random reasoning QA] [task instruction] [robot action] [image tokens] [task-completion verification QA] $\cdots$''.
    The task instruction is the  prompt used in existing VLA models, \emph{i.e.}, ``What should the robot do to finish the task \{task instruction\}'', where \{task instruction\} is the original task label from robot control data.
\end{itemize}

With these flexible interleaved formats, we sample the first, last, one random middle image from each segmented subtask video clips and annotate next subtask planning to construct interleaved temporal reasoning data. The interleaved spatial reasoning data is constructed using all robot trajectory prediction QA in embodied spatial reasoning data. The interleaved free chatting format randomly remaining QA data to connect embodied reasoning QA data and robot control data. Totally, we curated 122 thousand interleaved vision-text-action data for mixed-modality generation training.


\begin{figure}[t]     
    \includegraphics[width=1.0\textwidth]{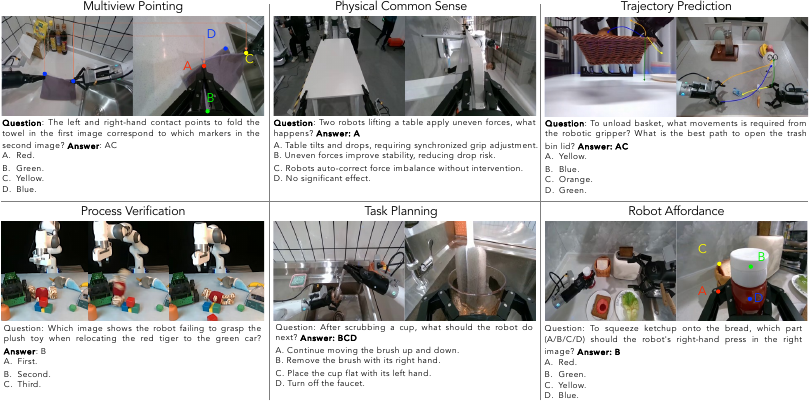}
    \caption{Examples of \benchmark, including Multiview Pointing, Physical Common Sense, Trajectory Prediction, Process Verification, Task Planning, and Robot Affordance.}
    \label{fig:benchmark_example}
\end{figure}

\subsection{Embodied Onevision Reasoning Benchmark}
Embodied Onevision Benchmark (\benchmark) aims to construct a comprehensive and balanced evaluation suite for open-world embodied reasoning, covering both challenging and accessible tasks. Existing benchmarks often suffer from the drawback that a single QA instance may conflate multiple reasoning aspects, such as combining spatial trajectories with extensive commonsense knowledge. This design leads to ambiguous evaluations, making it difficult to attribute performance to specific capabilities. In contrast, our benchmark ensures that each question targets a clearly defined reasoning angle rather than mixing multiple dimensions. This enables a more faithful diagnosis of models' strengths and weaknesses in an interpretable and disentangled manner.

As shown in~\cref{fig:statistic}(a), \benchmark is organized into four major categories: spatial understanding, physical commonsense, task reasoning, and state estimation. In total, the benchmark comprises 648 QA pairs manually labeled on diverse robot control data, distributed across categories as follows: 370 for spatial understanding, 140 for task reasoning, 84 for physical dynamic reasoning, and 48 for physical commonsense. This distribution reflects our intention to emphasize spatial reasoning, which is an essential component and the bottleneck of embodied intelligence. Together, these four categories provide a structured yet broad evaluation of a model’s ability to reason about space, physics, tasks, and states in embodied intelligence.

Each major category is further divided into sub-categories, as illustrated in \cref{fig:statistic,fig:benchmark_example}.
\begin{itemize}[leftmargin=*, topsep=0pt, noitemsep]
    \item \textbf{Spatial Understanding} covers trajectory reasoning (predicting object motion), visual grounding (localizing objects in context), relation reasoning (understanding relative positions among objects), and multiview pointing (identifying objects consistently across multiple viewpoints).
    \item \textbf{Physical Commonsense} includes direct influence and counterfactual reasoning, both of which require models to apply fundamental physical principles in embodied interactions.
    \item \textbf{Task Reasoning} spans task planning (deriving action sequences toward goals), episode captioning (summarizing task-level events), and process verification (evaluating task completion and progress).
    \item \textbf{State Estimation} involves recognizing object states (\emph{e.g.}, open/closed, full/empty), reasoning about robot-object interactions, and assessing action-level consequences.
\end{itemize}

By structuring the benchmark into these categories and sub-categories, we provide a principled way to evaluate embodied models across complementary aspects of embodied intelligence, while maintaining interpretability of each evaluation dimension.

\section{Experimental Results}
\label{sec:experiment}

\subsection{Implementation Details}
The \ours model is trained on a large-scale corpus that integrates 1.2M real-robot demonstrations from AgiBotWorld~\citep{bu2025agibot}, Open X-Embodiment~\citep{o2024open}, RoboMIND~\citep{wu2024robomind}, and SO100-Community~\citep{shukor2025smolvla}, together with 5.7M web multimodal samples and 1.5M interleaved embodied data pairs, yielding a total of 135B tokens. To optimize on this scale, we train for five epochs with Flash-Attention variable-length packing (average sequence length of 16,384) and a batch size of 1. The backbone learning rates are set to $5\times10^{-5}$ for both the language model and multimodal projector, $1\times10^{-5}$ for the vision encoder, and a fixed balance factor of $1.0$ is applied between language regression and action flow matching losses. Additional settings include an action denoising chunk size of $16$, resolution bounds of 50176–100352, and the DeepSpeed ZeRO-1 optimizer. At inference time, the policy is conditioned on multi-view camera observations and sub-task instructions. It predicts a 16-step action chunk through 10 denoising iterations for robot execution. This design achieves a balance between stability and efficiency, while requiring only 6 GB of GPU memory on a single NVIDIA RTX 4090, thus enabling real-time evaluation in both simulation and real-world environments.

\subsection{Open-sourced Benchmark Evaluation Setups}
To thoroughly evaluate the embodied reasoning and dexterous manipulation capabilities, we assess our model across two key categories of benchmarks:
\textit{Embodied Reasoning} and \textit{Robot Control}. They encompass a broad spectrum of perceptual, cognitive, and control challenges, spanning from high-level visual reasoning to low-level motor skill execution, across both simulated and visually complex real-world environments.

\noindent\textbf{Embodied Reasoning.} This category consists of three reasoning-centric benchmarks: \textbf{RoboVQA}, \textbf{ERQA}, and our self-constructed \benchmark. They are described in the following items. Specifically, these benchmarks are designed to evaluate distinct and complementary cognitive capabilities. To ensure standardized and consistent evaluation, we employ VLMEvalKit, a unified framework that facilitates rigorous benchmarking and reliable performance comparison across these tasks.

\begin{itemize}[leftmargin=*, topsep=0pt, noitemsep]
    \item \textbf{RoboVQA}~\citep{sermanet2024robovqa}: A free-form visual question answering (VQA) benchmark designed to assess high-level reasoning in egocentric embodied scenes. Following prior works, model performance is evaluated using BLEU-4 scores.
    \item \textbf{ERQA}~\citep{gemini_robotics}: A curated benchmark that emphasizes spatial reasoning and grounded world knowledge. The tasks are designed around realistic robotic scenarios requiring inference over object relationships, geometry, and agent-centric spatial cues.
    \item \textbf{\benchmark}: We propose \benchmark to evaluate models in \textit{spatial-temporal reasoning}. The benchmark consists of 700 multiple-choice VQA tasks, distributed evenly across four reasoning categories: \textbf{Physical Commonsense}, \textbf{Spatial Understanding}, \textbf{State Estimation}, and \textbf{Task Reasoning}. All tasks are grounded on egocentric visual inputs, meticulously constructed to reflect the real-world challenges of robot perception and decision-making.
\end{itemize}

\noindent\textbf{Robot Control.} This category includes two manipulation benchmarks: \textbf{SimplerEnv} and \textbf{LIBERO}, which collectively assess a wide spectrum of control capabilities across visually diverse, long-horizon, and goal-conditioned tasks.
\begin{itemize}[leftmargin=*, topsep=0pt, noitemsep]
    \item \textbf{SimplerEnv}~\citep{li24simpler}: A manipulation benchmark featuring WidowX and Google Robots operating in visually diverse environments with variations in lighting, surface textures, and camera viewpoints. It leverages the Bridge Data V2\citep{walke2023bridgedata} for evaluating Real-to-Sim transfer. SimplerEnv includes both short-horizon atomic tasks and visually challenging variants, specifically designed to assess robustness under appearance shifts between real and simulated settings.
    \item \textbf{LIBERO}~\citep{liu2023libero}: A long-horizon manipulation benchmark consisting of temporally extended, multi-stage tasks across various object categories and interaction types. LIBERO emphasizes planning, memory, and semantic understanding in complex simulated environments.
\end{itemize}

\subsection{Benchmark Evaluation Results}
\noindent\textbf{Embodied Reasoning.} To evaluate the embodied reasoning capabilities, we benchmark it on three key tasks: \textbf{RoboVQA}~\cite{sermanet2024robovqa}, \textbf{ERQA}~\citep{gemini_robotics}, and \benchmark. These benchmarks assess a diverse range of reasoning skills, including spatial understanding, physical commonsense, and multistep task planning. We compare \ours against three categories of baselines: \textit{public VLMs} (\emph{e.g.}, Qwen2.5 VL~\citep{qwen2_5_vl}, InternVL2.5~\citep{internvl}), \textit{private VLMs} (\emph{e.g.}, GPT-4o, Gemini 1.5 Flash~\citep{gemini}, Claude 3.5), and \textit{co-trained visual-language-action models} (\emph{e.g.}, ChatVLA~\citep{zhou2025chatvla}, Magma~\citep{yang2025magma}), with all models evaluated under consistent embodied conditions. 
\begin{table*}[t]
    \centering
    \resizebox{1.0\textwidth}{!}{
        \begin{tabular}{lcccccccc}
            \toprule
            \bf Multi-modal Benchmark               &  & \bf RoboVQA &  & \bf ERQA & & \bf \benchmark@ Spatial & \bf \benchmark@ Temporal & \bf Overall \\
            \midrule
            Gemini 1.5 Flash~\citep{team2023gemini} &  & 46.0        &  & 46.3     & & 46.3                & 37.8            & 44.1        \\
            Claude 3.5                              &  & 26.7        &  & 35.5     & & 24.0                & 34.8            & 30.3        \\
            GPT-4o 2024-11-20                       &  & 47.2        &  & 40.0     & & 35.6                & 39.3            & 40.5        \\
            \midrule
            Qwen2.5 VL 3B~\citep{qwen2_5_vl}        &  & 55.9        &  & 35.3     & & 20.0                & 22.6            & 33.5        \\
            Qwen2.5 VL 7B~\citep{qwen2_5_vl}        &  & 56.9        &  & 39.3     & & 26.7                & 31.5            & 38.6        \\
            InternVL2.5 4B~\citep{internvl2_5}      &  & 45.7        &  & 36.8     & & 37.0                & 31.9            & 37.9        \\
            InternVL2.5 8B~\citep{internvl2_5}      &  & 43.9        &  & 45.2     & & 32.8                & 38.1            & 40.0        \\
            \midrule
            ChatVLA 2B~\citep{zhou2025chatvla}      &  & 33.7        &  & 34.3     & & 26.4                & 21.9            & 29.1        \\
            Magma 8B~\citep{yang2025magma}          &  & 30.3        &  & 29.3     & & 29.4                & 36.7            & 31.4        \\
            \rowhighlight
            \ours (3B)                              &  & 58.5        &  & 45.5     & & 36.4                & 38.9            & 44.8        \\
            \bottomrule
        \end{tabular}
    }
    \caption{Performance comparison with standard VLMs and co-trained visual-language-action models on \textbf{RoboVQA}~\citep{sermanet2024robovqa}, \textbf{ERQA}~\citep{gemini_robotics}, and \benchmark benchmarks.}
    \label{tab:multimodal}
\end{table*}
Across all the benchmarks presented in~\cref{tab:multimodal}, \ours{} consistently demonstrates robust reasoning capabilities in embodied scenarios in the open world. On \textbf{RoboVQA}, which evaluates long-horizon visuospatial inference in egocentric settings, \ours{} achieves the BLEU-4 score of 58.5, significantly outperforming leading closed-source models, such as GPT-4o (47.2). For \textbf{ERQA}, a spatially grounded benchmark derived from real-world visual scenes, \ours{} achieves the accuracy of 45.5, surpassing InternVL2.5 8B (45.2) and outperforming other public and private VLMs. In the suite \benchmark introduced in this paper, mainstream VLMs achieve a modest average score of 32, highlighting the substantial limitations of current methods in handling embodied tasks. On the \textbf{\benchmark@~Spatial} leaderboard, which evaluates performance across multiview reasoning, trajectory prediction, and visual grounding tasks, \ours{} outperforms other open-source models of comparable size, securing a score of 36.4. This highlights \ours{}'s improved capability in multi-modal scene understanding and spatial reasoning. Furthermore, in the \textbf{\benchmark@~Temporal} subset, dedicated to multimodal and temporal reasoning in manipulation tasks, \ours{} achieves 38.9, underscoring its strong aptitude for scene-level abstraction and temporal prediction. 
\begin{table}[t]
    \centering
    \resizebox{1.0\textwidth}{!}{%
        \begin{tabular}{lcccccccccc}
            \toprule
            \multirow{2}{*}{\bf Model}                      & \multicolumn{2}{c}{\bf LIBERO-Spatial} & \multicolumn{2}{c}{\bf LIBERO-Object} & \multicolumn{2}{c}{\bf LIBERO-Goal} & \multicolumn{2}{c}{\bf LIBERO-Long} & \multicolumn{2}{c}{\bf Overall}                                                                              \\
                                                            & SR                                     & Rank                                  & SR                                  & Rank                                & SR                              & Rank & SR                        & Rank & SR                        & Rank \\
            \midrule
            OpenVLA~\citep{kim2024openvla}                  & 84.7 $\pm$ 0.9\%                       & 5                                     & 88.4 $\pm$ 0.8\%                    & 5                                   & 79.2 $\pm$ 1.0\%                & 4    & 53.7 $\pm$ 1.3\%          & 5    & 76.5 $\pm$ 0.6\%          & 5    \\
            SpatialVLA~\citep{qu2025spatialvla}             & 88.2 $\pm$ 0.5\%                       & 2                                     & 89.9 $\pm$ 0.7\%                    & 4                                   & 78.6 $\pm$ 0.6\%                & 5    & 55.5 $\pm$ 1.0\%          & 4    & 78.1 $\pm$ 0.7\%          & 4    \\
            OpenVLA-OFT~\citep{kim2025fine}                 & \textbf{97.6 $\pm$ 0.9\%}              & 1                                     & \textbf{98.4 $\pm$ 0.8\%}           & 1                                   & \textbf{97.9 $\pm$ 1.0\%}       & 1    & \textbf{94.5 $\pm$ 1.3\%} & 1    & \textbf{97.1 $\pm$ 0.6\%} & 1    \\
            ThinkAct~\citep{huang2025thinkact}              & 88.3 $\pm$ --\%                        & 3                                     & 91.4 $\pm$ --\%                     & 3                                   & 87.1 $\pm$ --\%                 & 3    & 70.9 $\pm$ --\%           & 3    & 84.4 $\pm$ --\%           & 3    \\
            MolmoAct-7B-D~\citep{lee2025molmoact}           & 87.0 $\pm$ --\%                        & 4                                     & 95.4 $\pm$ --\%                     & 2                                   & 87.6 $\pm$ --\%                 & 2    & 77.2 $\pm$ --\%           & 2    & 86.6 $\pm$ --\%           & 2    \\
            \midrule
            Diffusion Policy~\citep{chi2024diffusionpolicy} & 78.5 $\pm$ 1.1\%                       & 6                                     & 87.5$\pm$ 0.7\%                     & 5                                   & 73.5$\pm$ 1.2\%                 & 6    & 64.8$\pm$ 1.3\%           & 4    & 76.1 $\pm$ 0.7\%          & 5    \\
            Octo~\citep{team2024octo}                       & 78.9 $\pm$ 1.0\%                       & 5                                     & 85.7 $\pm$ 0.9\%                    & 6                                   & 84.6 $\pm$ 0.9\%                & 5    & 51.1 $\pm$ 1.3\%          & 6    & 75.1 $\pm$ 0.6\%          & 6    \\
            $\pi_{0}$~\citep{black2024pi_0}                 & 96.8 $\pm$ 0.8\%                       & 2                                     & 98.8 $\pm$ 0.9\%                    & 2                                   & 95.8 $\pm$ 1.1\%                & 2    & 85.2 $\pm$ 1.2\%          & 3    & 94.2 $\pm$ 0.9\%          & 2    \\
            $\pi_{0}$-FAST~\citep{pertsch2025fast}          & 96.4 $\pm$ 0.7\%                       & 3                                     & 96.8 $\pm$ 0.7\%                    & 4                                   & 88.6 $\pm$ 1.0\%                & 4    & 60.2 $\pm$ 1.4\%          & 5    & 85.5 $\pm$ 1.0\%          & 4    \\
            GR00T N1~\citep{bjorck2025gr00t}                & 94.4 $\pm$ 0.9\%                       & 4                                     & 97.6 $\pm$ 1.0\%                    & 3                                   & 93.0$\pm$ 1.2\%                 & 3    & 90.6$\pm$ 1.0\%           & 2    & 93.9 $\pm$ 1.1\%          & 3    \\
            \rowhighlight
            \ours~(Ours)                                            & \textbf{99.7 $\pm$ 0.2\%}              & 1                                     & \textbf{99.8 $\pm$ 0.1\%}           & 1                                   & \textbf{99.2 $\pm$ 0.3\%}       & 1    & \textbf{94.8 $\pm$ 0.6}\% & 1    & \textbf{98.2 $\pm$ 0.3\%} & 1    \\
            \bottomrule
        \end{tabular}}
    \vspace{-0.5em}
    \caption{Performance comparison with state-of-the-art policies on \textbf{LIBERO Benchmark}~\citep{liu2023libero}.}
    \label{tab:libero}
\end{table}
\begin{table*}[!h]
    \hspace*{-0.0075\textwidth}
    \centering
    \vspace{-2ex}
    \resizebox{1.005\textwidth}{!}{
        \begin{tabular}{lccccc}
            \toprule
            \bf WidowX Benchmark                     & \bf Put Spoon on Towel & \bf Put Carrot on Plate & \bf Stack Blocks            & \bf Put Eggplant in Basket & \bf Overall \\
            \midrule
            RT-1-X~\citep{o2024open}                 & 0\%                    & 4.2\%                   & 0\%                         & 0\%                        & 1.1\%       \\
            OpenVLA~\citep{kim2024openvla}           & 0\%                    & 0\%                     & 0\%                         & 4.1\%                      & 1.0\%       \\
            SpatialVLA~\citep{qu2025spatialvla}      & 20.8\%                 & 20.8\%                  & 25.0\%                      & 70.8\%                     & 34.4\%      \\
            Magma~\citep{yang2025magma}              & 37.5\%                 & 29.2\%                  & 20.8\%                      & 91.7\%                     & 44.8\%      \\
            $\pi_{0}$-FAST~\citep{pertsch2025fast}   & 29.1\%                 & 21.9\%                  & 10.8\%                      & 66.6\%                     & 32.1\%      \\
            \midrule
            Octo-Base~\citep{team2024octo}           & 12.5\%                 & 8.3\%                   & 0\%                         & 43.1\%                     & 16.0\%      \\
            Octo-Small~\citep{team2024octo}          & 47.2\%                 & 9.7\%                   & 4.2\%                       & 56.9\%                     & 29.5\%      \\
            RoboVLM~\citep{li2023generalist}         & 45.8\%                 & 20.8\%                  & 4.2\%                       & 79.2\%                     & 37.5\%      \\
            $\pi_{0}$~\citep{black2024pi_0}          & 83.8\%                 & 52.5\%                  & 52.5\%                      & 87.9\%                     & 69.2\%     \\
            ThinkAct~\citep{huang2025thinkact}       & 58.3\%                 & 37.5\%                  & 8.7\%                       & 70.8\%                     & 43.8\%     \\
            \rowhighlight
            \ours (Ours)                             & 63.6\%                 & 54.5\%                  & 81.8\%                      & 90.9\%                     & \bf 72.7\%  \\
            \midrule
            \bf Google Robot Benchmark (Matching)    & \bf Pick Coke Can      & \bf Move Near           & \bf Open$\cdot$Close Drawer & \bf Drawer Apple           & \bf Average \\
            \midrule
            RT-1~\citep{brohan2022rt}                & 85.7\%                 & 44.2\%                  & \textbf{73.0\%}             & 6.5\%                      & 52.4\%      \\
            OpenVLA~\citep{kim2024openvla}           & 16.3\%                 & 46.2\%                  & 35.6\%                      & 0\%                        & 24.5\%      \\
            TraceVLA~\citep{zheng2024tracevla}       & 28.0\%                 & 53.7\%                  & 57.0\%                      & 0\%                        & 34.7\%      \\
            SpatialVLA~\citep{qu2025spatialvla}      & 86.0\%                 & 77.9\%                  & 57.4\%                      & 0\%                        & 55.3\%      \\
            Magma~\citep{yang2025magma}              & 75.0\%                 & 53.0\%                  & 58.9\%                      & 8.3\%                      & 48.8\%      \\
            $\pi_{0}$-FAST~\citep{pertsch2025fast}   & 75.3\%                 & 67.5\%                  & 42.9\%                      & 0                          & 46.4\%      \\
            \midrule
            RT-1-X~\citep{o2024open}                 & 56.7\%                 & 31.7\%                  & 59.7\%                      & 40.7\%                     & 47.2\%      \\
            RT-2-X~\citep{o2024open}                 & 78.7\%                 & 77.9\%                  & 25.0\%                      & 7.4\%                      & 47.3\%      \\
            Octo-Base~\citep{team2024octo}           & 17.0\%                 & 4.2\%                   & 22.7\%                      & 0                          & 11.0\%      \\
            RoboVLM~\cite{li2023generalist}          & 77.3\%                 & 61.7\%                  & 43.5\%                      & 24.1\%                     & 51.7\%      \\
            $\pi_{0}$~\citep{black2024pi_0}          & 97.9\%                 & 78.7\%                  & 62.25\%                     & 46.6\%                     & 71.4\%      \\
            ThinkAct~\citep{huang2025thinkact}       & 92.0\%                 & 72.4\%                  & 50.0\%                      & --                         & --          \\
            MolmoAct~\citep{lee2025molmoact}         & 77.7\%                 & 77.1\%                  & 60.0\%                      & --                         & --          \\
            \rowhighlight
            \ours (Ours)                             & 98.0\%                 & 83.8\%                  & 71.3\%                      & 52.8\%                     & 76.5\%      \\
            \midrule
            \bf Google Robot Benchmark (Aggregation) & \bf Pick Coke Can      & \bf Move Near           & \bf Open$\cdot$Close Drawer & \bf Drawer Apple           & \bf Average \\
            \midrule
            RT-1~\citep{brohan2022rt}                & 89.8\%                 & 50.0\%                  & 32.3\%                      & 2.6\%                      & 43.7\%      \\
            OpenVLA~\citep{kim2024openvla}           & 54.5\%                 & 47.7\%                  & 17.7\%                      & 0.0\%                      & 30.0\%      \\
            TraceVLA~\citep{zheng2024tracevla}       & 60.0\%                 & 56.4\%                  & 31.0\%                      & 0\%                        & 36.9\%      \\
            SpatialVLA~\citep{qu2025spatialvla}      & 88.0\%                 & 72.7\%                  & 41.8\%                      & 6.3\%                      & 52.2\%      \\\
            Magma~\citep{yang2025magma}              & 68.6\%                 & 78.5\%                  & 59.0\%                      & 24.0\%                     & 57.5\%      \\
            $\pi_{0}$-FAST~\citep{pertsch2025fast}   & 77.6\%                 & 68.2\%                  & 31.3\%                      & 0                          & 44.3\%      \\
            \midrule
            RT-1-X~\citep{o2024open}                 & 49.0\%                 & 32.3\%                  & 29.4\%                      & 10.1\%                     & 30.2\%      \\
            RT-2-X~\citep{o2024open}                 & 82.3\%                 & 79.2\%                  & 35.3\%                      & 20.6\%                     & 54.4\%      \\
            Octo-Base~\citep{team2024octo}           & 0.6\%                  & 3.1\%                   & 1.1\%                       & 0                          & 1.2\%       \\
            RoboVLM~\cite{li2023generalist}          & 75.6\%                 & 60.0\%                  & 10.6\%                      & 0                          & 36.6\%      \\
            $\pi_{0}$~\citep{black2024pi_0}          & 90.1\%                 & 80.7\%                  & 27.6\%                      & 20.5\%                     & 54.7\%      \\
            ThinkAct~\citep{huang2025thinkact}       & 84.0\%                 & 63.8\%                  & 47.6\%                      & --                         & --          \\
            MolmoAct                                 & 76.1\%                 & 61.3\%                  & 78.8\%                      & --                         & --          \\
            \rowhighlight
            \ours (Ours)                             & 91.6\%                 & 81.7\%                  & 55.0\%                      & 23.8\%                     & 63.0\%      \\
            \bottomrule
        \end{tabular}
    }
    \caption{Performance comparison with state-of-the-art policies on \textbf{SimplerEnv Benchmark}~\citep{li24simpler}.}
    \label{tab:benchmark_simpler}
    \vspace{-2ex}
\end{table*}

\noindent\textbf{Robot Control.} We evaluated manipulation performance on two simulation benchmarks: \textbf{LIBERO}~\citep{liu2023libero}and \textbf{SimplerEnv}~\citep{li24simpler}. LIBERO spans four subsets, Spatial, Object, Goal, and Long, designed to test generalization across spatial layouts, object categories, goal semantics, and long-horizon planning. SimplerEnv includes Google-VM (Visual Matching), Google-VA (Visual Aggregation), and WindoWX setups with controlled variations in color, texture, lighting, and camera pose, assessing robustness under visual distribution shift.  We compare against leading robot foundation models, including the \textbf{RT series}~\citep{brohan2022rt,o2024open}, \textbf{Octo}~\citep{team2024octo}, \textbf{OpenVLA}~\citep{kim2024openvla}, the \boldmath$\pi$\unboldmath-series~\citep{black2024pi_0,pertsch2025fast}, and \textbf{ThinkAct}~\citep{huang2025thinkact}, among others.

As shown in~\cref{tab:libero} and ~\cref{tab:benchmark_simpler}, our method achieves superior performance across various robotic manipulation tasks. On the \textbf{LIBERO} benchmark, it attains an average success rate of 98.2\%, outperforming recent state-of-the-art models, including OpenVLA-OFT~\citep{kim2025fine}, $\pi_0$~\citep{black2024pi_0}, and GR00t-N1~\citep{bjorck2025gr00t} by 1.1\%, 4.0\%, and 4.3\%, respectively, demonstrating robust generalization across spatial, semantic, and long-horizon tasks. On the \textbf{SimplerEnv} benchmarks, it surpasses $\pi_0$~\citep{black2024pi_0} by 3.5\%, 5.1\%, and 8.3\% in the WidowX, Google-VM, and Google-VA variants, respectively, achieving the highest success rates of 72.7\%, 76.5\%, and 63.0\%. These results are obtained with lightweight fine-tuning on a modest mixed-modality dataset, underscoring the method’s data efficiency, dexterous control, and precise visuomotor grounding.

\subsection{Real-world Experiment Evaluation Setups}
We introduce a comprehensive set of robotic manipulation tasks to evaluate the generalist capabilities of the VLA model across a diverse array of real-world scenarios. These tasks span pick-and-place, articulated object manipulation, long-horizon and dexterous actions, while also challenging the robots' abilities to perform embodied reasoning, engage in complex planning, and adapt to varied control types, as illustrated in~\cref{fig:task_example}. These tasks are described as follows.
\begin{figure*}[htbp]
    \centering
    \vspace{-2ex}
    \includegraphics[width=1\linewidth]{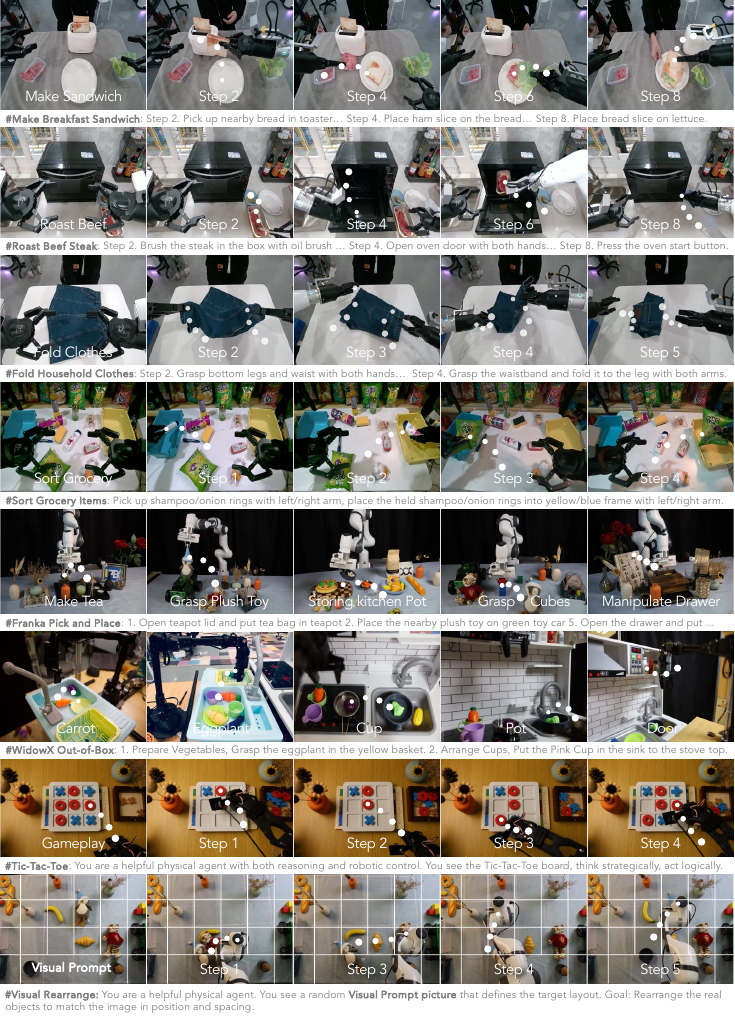}
    \vspace{-4ex}
    \caption{Example of real-world evaluation tasks in diverse robots, including Agibot G-1 Long-horizon Dexterous (row 1-4), Franka Panda Pick-and-Place (row 5), WidowX 250 S Out-of-Box (row 6), and Embodied Reasoning Control in Franka, Agibot G-1, and Lerobot SO100 (row 1,2,7,8).}
    \label{fig:task_example}
\end{figure*}

\begin{enumerate}[leftmargin=*, topsep=0pt, noitemsep]
    \item \textbf{Franka Panda Pick-and-Place (7 Tasks).}  Household tasks such as brewing tea, storing kitchen items, and sorting cubes, demonstrating proficiency in object-to-container pick-and-place and articulated object manipulation. These tasks test fine motor control and adaptability to diverse object geometries in basic household settings. 

    \item \textbf{WidowX 250 S Out-of-Box (13 Tasks).} Out-of-box evaluation in a kitchen environment includes tasks involving both object-to-container manipulation (\emph{e.g.}, vegetable preparation, cup arrangement) and articulated handling (\emph{e.g.}, door closing), highlighting the robot’s flexibility in managing diverse objects and complex actions.

    \item \textbf{Agibot G-1 Long-Horizon Dexterity (4 Tasks).}  Long-horizon, dexterous tasks requiring sequential planning and fine manipulation, such as folding clothes, making sandwiches, and sorting groceries. These tasks highlight the robot’s proficiency in multi-step operations in multi-step tasks, demonstrating advanced dexterity and adaptability in long-term execution scenarios.

    \item \textbf{Embodied Reasoning Control (4 Tasks).} Physical and abstract tasks including visual object rearrangement, tic-tac-toe, and long-horizon planning tasks (\emph{e.g.}, making breakfast sandwich, roasting beef steak), showcasing its embodied reasoning capabilities. These tasks require not only precise manipulation but also higher-level reasoning, testing the intersection of manipulation and reasoning in real-world contexts.
\end{enumerate}
These 28 task settings serve to assess the model’s performance in four key dimensions: mastering diverse manipulations across multiple robot platforms, specializing in long-horizon dexterous tasks, emerging open-world embodied generalization, and enhanced generalization through unified reasoning capabilities. The results of the four dimensions are presented successively in the following subsections.
\subsection{Mastering Diverse Manipulations on Multiple Embodiments}
\begin{figure*}[htbp]
    \centering
    \hspace*{-0.05\textwidth}
    \includegraphics[width=1.06\linewidth]{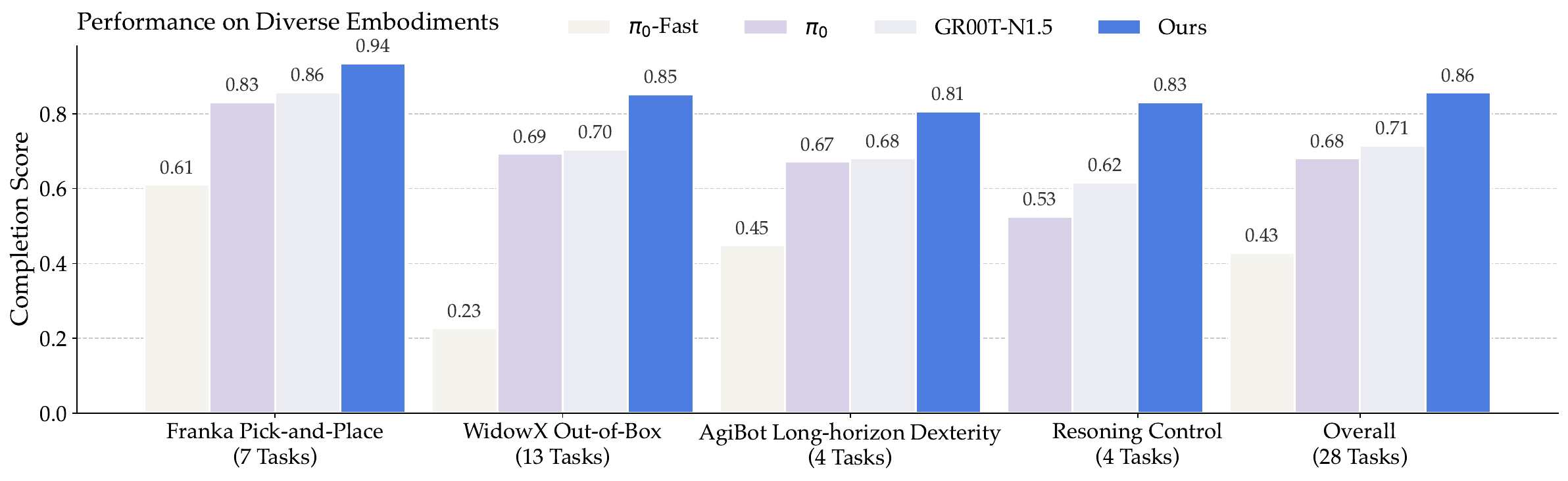}
    \caption{Performance comparison on diverse robot platforms and task categories.}
    \label{fig:real_diverse}
\end{figure*}
In this section, we demonstrate that \ours{} can perform a wide range of dexterous manipulation tasks across various robotic platforms, showcasing its robustness and adaptability. We evaluate its performance on both short-horizon and long-horizon tasks using different robots, including Franka Panda, WidowX 250 S, Agibot G-1, and Lerobot SO100. These tasks range from basic pick-and-place to complex articulated object manipulation and long-horizon tasks. We compare \ours{} against state-of-the-art baselines, including $\pi_0$-Fast, $\pi_0$, and GR00T-N1.5~\citep{pertsch2025fast, black2024pi_0, bjorck2025gr00t}, highlighting its superior performance across these varied tasks.

The results reported in~\cref{fig:real_diverse} show that \ours consistently outperforms the baselines across all robot platforms and task categories, achieving a performance of 86.0\%, compared with 43.0\% for Fast, 71.0\% for GR00T-N1.5, and 68.0\% for $\pi_0$. Notably, in long-horizon tasks with Agibot G-1, \ours achieves a remarkable completion score of 81.0\%, surpassing the performance of $\pi_0$ at 67.0\%. Similarly, on Franka Panda’s pick-and-place tasks, \ours scores 94.0\%, exceeding GR00T-N1.5’s 86.0\%. On more complex embodied reasoning control tasks, \emph{e.g.}, Tic-Tac-Toe and Visual Rearrangement, \ours maintains a completion score of 83.0\%, which again surpasses that of $\pi_0$ at 53.0\% and GR00T-N1.5 at 62.0\%. These results indicate that \ours excels at managing multi-step and dexterous tasks that demand high precision and adaptability across various environments and robot platforms. The strong performance of \ours across such a diverse set of tasks underscores its potential for real-world deployment, where robotic systems are required to operate reliably in dynamic and varied environments.
\subsection{Specializing to Long-horizon Dexterity}
\label{sec:dexterity}
In this section, we investigate the \ours model’s ability to specialize in long-horizon dexterous tasks, which require precise coordination and sustained execution over extended time periods. While the model demonstrated strong performance on short-horizon tasks in previous evaluations, these more complex tasks present unique challenges, as they involve a sequence of coordinated actions across multiple steps. To explore the model's specialization potential, we fine-tune \ours using high-quality demonstration data for several real-world tasks that push the limits of both dexterous and long-horizon manipulation. As illustrated in~\cref{fig:task_example} (row 1-4), we select four tasks that require intricate multi-step decisions and fine manipulation across multiple sequential actions:
\begin{enumerate}[leftmargin=*, topsep=0pt, noitemsep]
    \item \textbf{Make Breakfast Sandwich}:
    Tasks involve the precise placement of bread, ham, lettuce, and other ingredients in a multi-step sequence using both arms to assemble the sandwich.
    
    \item \textbf{Roast Beef Steak}:
   Tasks Involve a series of steps requiring fine motor skills, including brushing oil on the steak, placing it in the oven, and operating the oven with both arms.
    
    \item \textbf{Fold Household Clothes}:
    Tasks involve grasping and folding shorts by coordinating both arms for precise control over the fabric, requiring dexterity and spatial awareness.
    
    \item \textbf{Sort Grocery Items}:
    Tasks involve sort items (\emph{e.g.}, shampoo and green onions) into designated containers based on specific instructions, testing the robot's instruction-following and object manipulation skills.
\end{enumerate}
\begin{figure*}[htbp]
    \centering
    \hspace*{-0.05\textwidth}
    \includegraphics[width=1.06\linewidth]{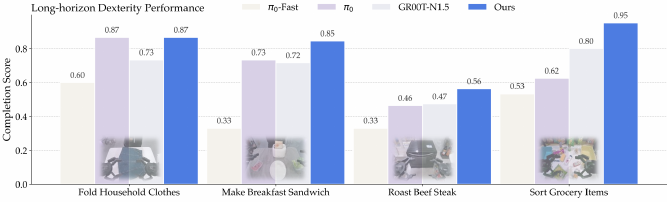}
    \caption{Long-horizon dexterity completion rate comparison on the AgiBot G-1 platform.}
    \label{fig:dexterity}
\end{figure*}
Specializing fine-tuning is performed using 150 demonstration trajectories per task, with each suite comprising approximately 0.5 million frames across 8 steps. The training process spans 25 epochs, with a 9:1 sub-task-to-overall task mixture ratio, where 90\% of the data is from sub-tasks and 10\% from the overall task. During evaluation, 10 test runs per task are conducted, using sub-task instructions to compute the average completion score.

As shown in~\cref{fig:dexterity}, we assess the specialized \ours model on four long-horizon tasks, comparing its performance against leading methods. Specifically, on the Roast Beef Steak task, which requires fine motor control and interaction with kitchen appliances, \ours reaches a completion score of 56.0\%, substantially higher than both GR00T-N1.5 (47.0\%) and $\pi_0$ (46.0\%). On the Make Breakfast Sandwich task, \ours demonstrates an 85.0\% completion score, far surpassing $\pi_0$ (73.0\%) and GR00T-N1.5 (72.0\%), showcasing its ability to handle complex, multi-step manipulations. Similarly, in the Sorting Grocery Items task, which tests both object manipulation and instruction-following abilities, \ours excels with a completion score of 95.0\%, while $\pi_0$ and GR00T-N1.5 achieve notably lower success rates (62.0\% and 80.0\%, respectively). Overall, as shown in~\cref{fig:real_diverse}(c), \ours achieves an impressive average completion score of 81.0\%, significantly outperforming both GR00T-N1.5 (68.0\%) and $\pi_0$ (67.0\%). These results highlight \ours' ability to excel in long-horizon tasks, showing significant improvements over baseline models, and its capacity to manage complex, multi-step actions in real-world scenarios. By seamlessly integrating regression and flow matching within a unified backbone, the model demonstrates the potential for more stable inferences and precise decision-making, especially in long-sequence tasks, without relying on action-specific parameters or introducing heuristic bottlenecks.

\subsection{Emerging Open-world Embodied Generalization}
\label{sec:generalization}
\begin{figure*}[t]
    \centering
    \includegraphics[trim=2 0 0 0,width=1.0\linewidth]{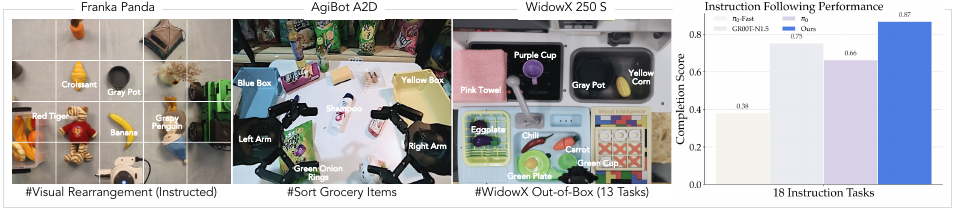}
    \caption{\textbf{Instruction-following in open-world settings.} Left: Example scenes from three task types: \textbf{Instructed Visual Rearrangement} (Franka Panda), \textbf{Sort Grocery Items} (Agibot G-1), and \textbf{WidowX Out-of-Box}. Right: Success rates over 18 tasks.}
    \label{fig:instruction-following}
\end{figure*}
The key challenge for embodied foundation models is generalizing to open-world scenarios where natural language instructions must be grounded into precise, executable actions. To evaluate this capability, we perform a generalization assessment using a suite of 15 instruction-following tasks across three different embodiments: \textbf{Franka Panda}, \textbf{Agibot G-1}, and \textbf{WidowX 250 S}, all operating in unstructured, object-rich environments. The tasks include \textbf{Visual Rearrangement (Instructed)}, which involves 7 instructions over 5 objects to test spatial instruction following and sequential manipulation; \textbf{Sort Grocery Items}, requiring fine-grained recognition and placement of 4 objects under cluttered conditions; and \textbf{WidowX Out-of-Box}, which consists of 13 multi-instruction tasks involving both rigid and articulated objects, assessing adaptability in compact workspaces.

Against strong baselines GR00T-N1.5 and $\pi_0$~\citep{pertsch2025fast,black2024pi_0,bjorck2025gr00t}, \ours{} achieves an average completion score of \textbf{87.0\%}, outperforming GR00T-N1.5 (75.0\%) and $\pi_0$ (66.0\%). Performance gains are consistent across embodiments and task types, with \ours{} maintaining high accuracy even under precise spatial references and unseen object–instruction combinations, demonstrating robustness to both language and visual distribution shifts. We observe that while $\pi_0$ executes tasks quickly and accurately, it struggles to follow instructions, often prioritizing objects closer to itself instead of adhering to the specified commands. This behavior may be due to the negative impact of pre-training on the entire backbone, which can lead to suboptimal decision-making in dynamic environments. In contrast, GR00T-N1.5, which freezes the language backbone during training, shows improved instruction-following capabilities.



\begin{figure*}[!ht]
    \centering
    \begin{subfigure}{1\linewidth}
        \centering
        \includegraphics[width=1.0\linewidth]{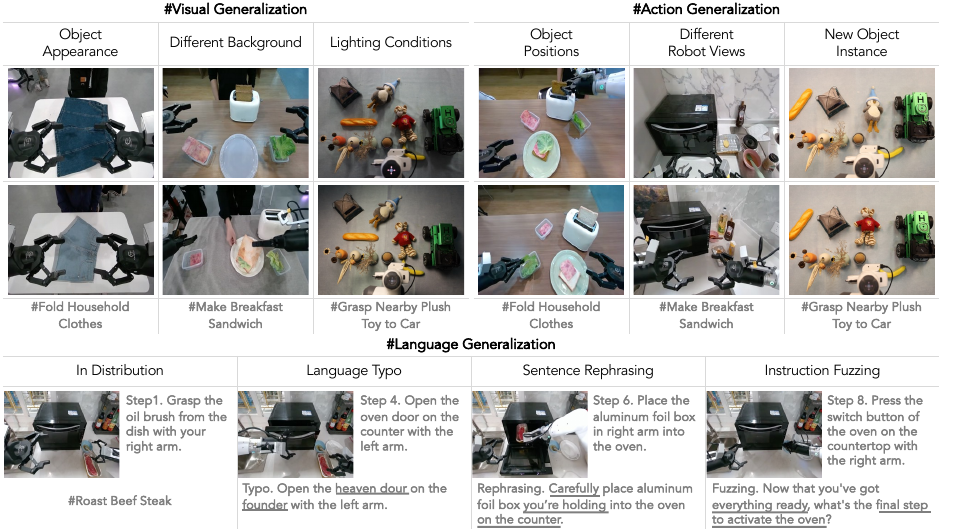}
        \caption{\textbf{Examples of the three generalization axes.} \textbf{Visual:} Changes in object appearance, background, and lighting. \textbf{Action:} Altered object positions, novel viewpoints, or new object instances. \textbf{Language:} Typos, rephrasings, and fuzzy descriptions.}
        \label{fig:generalization_example}
    \end{subfigure}
    \begin{subfigure}{1\linewidth}
        \centering
        \hspace{-5ex}
        \includegraphics[width=1.02\linewidth]{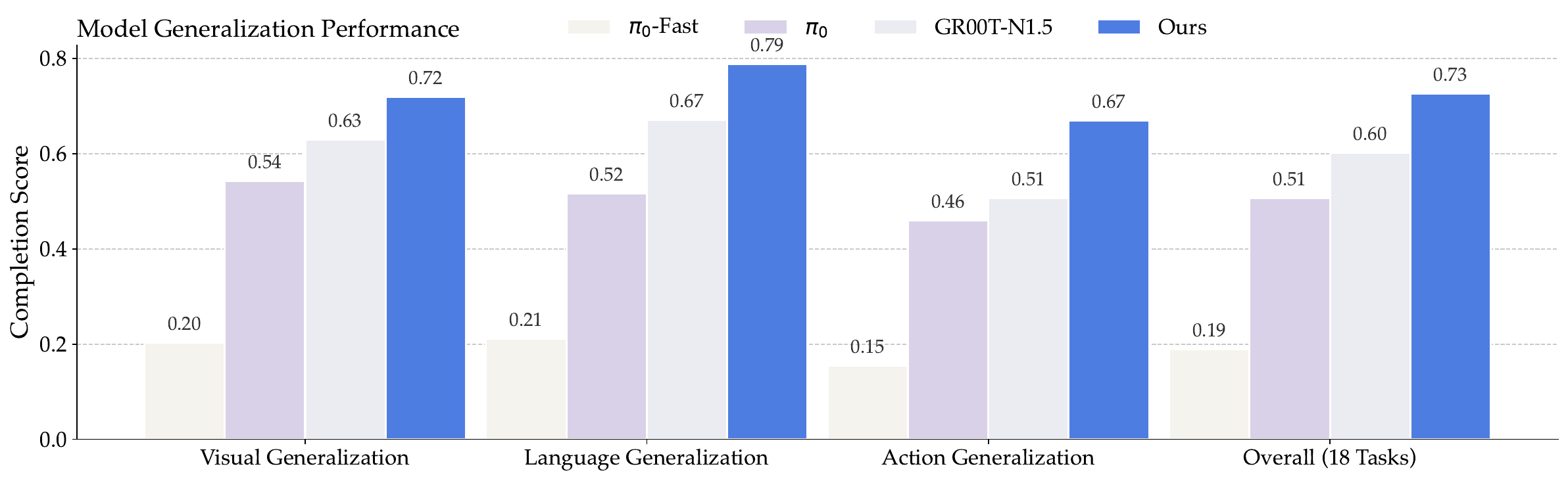}
        \caption{\textbf{Generalization performance breakdown.} Success rates for \ours{}, GR00T-N1.5, and $\pi_0$ across visual, action, and language variations.}
        \label{fig:bar_generalization}
    \end{subfigure}
    \caption{Generalization illustration and performance evaluation with visual, action, and language variations.}
    \label{fig:generalization_combined}
\end{figure*}

As shown in~\cref{fig:generalization_example}, we further extend the evaluation to \textbf{12 tasks} designed to probe generalization along three orthogonal variation axes--\textbf{Visual Generalization}: invariance to changes in object appearance, background, and lighting, tested on {WidowX Out-of-Box}, {Franka Panda Pick-and-Place}, and {Agibot G-1} household tasks ({Fold Clothes}, {Make Sandwich}, {Roast Beef}). \textbf{Action Generalization}: adaptation to altered object positions, new viewpoints, and unseen object instances with different physical properties. \textbf{Language Generalization}: robustness to typos, paraphrases, and vague or imprecise instruction formulations. The results are shown in Figure~\ref{fig:bar_generalization}, \ours{} attains the highest overall completion score (\textbf{73.0\%}) across all settings, outperforming GR00T-N1.5 (60.0\%) and $\pi_0$ (51.0\%). The advantage holds across all variation types: 67.0\% in \textbf{Action}, 79\% in \textbf{Language}, and 72.0\% in \textbf{Visual}. The largest relative gain appears in language variations, indicating strong linguistic robustness, which is a persistent weakness of many current VLM-based policies. These findings confirm that interleaved vision–text–action training not only strengthens in-distribution manipulation performance but also provides resilience to distribution shifts in perception, control, and instructions, which is critical for real-world deployment of generalist robot policies.

We observe that \ours demonstrates strong robustness \textbf{\textit{In Action Domain}} tasks, such as instruction variations like "Carefully place the aluminum foil box you’re holding into the oven on the counter" versus "Place the aluminum foil box in right arm into the oven.", or changes in lighting conditions. This is attributed to the model's co-training on interleaved vision-text-action datasets, including grounding boxes, 2D trajectories, instruction fuzzing, and task reasoning corpora. This multi-modal training approach enhances \ours' generalization ability in open-world scenarios, allowing it to adapt to diverse and dynamic environments. Despite these strengths, we also encounter challenges in generalizing \textbf{\textit{Out of the Action Domain}}, particularly when working with limited robot control data. For example, in the Make Breakfast Sandwich task, variations in the position of the bread lead to a significant drop in success rates. On the other hand, in the WidowX Out-of-Box task with large-scale Bridge data, where densely annotated interleaved data is available, we see a marked improvement in success rates. This suggests that interleaved vision-text-action co-training can fine-tune the model at the feature level, allowing it to map actions better and thus enhancing its generalization potential.

\subsection{Enhanced Generalization with Unified Reasoning}
To evaluate whether a \emph{sole} interleaved vision–text–action policy can seamlessly integrate high-level reasoning with low-level control in real environments, we design a reasoning-control benchmark comprising four tasks: \textbf{Visual Rearrangement}, \textbf{Tic-Tac-Toe}, \textbf{Make Breakfast Sandwich}, and \textbf{Roast Beef Steak}. These tasks require joint perception, spatial reasoning, multi-step planning, and bimanual manipulation under real-world constraints. Unlike hierarchical baselines, GR00T-N1.5 and $\pi_0$, which use an external LLM planner with a separate controller (GPT-4o), \ours{} integrates planning and control within a single decoder, aligning reasoning and motor execution in an interleaved sequence. The four tasks are described as follows.
\begin{enumerate}[leftmargin=*, topsep=0pt, noitemsep]
    \item \textbf{Visual Rearrangement.} The robot must arrange objects in a target layout according to a reference image (\emph{visual prompt}). It retrieves and places items sequentially, considering spatial relationships and maintaining collision-free motion. Success depends on interpreting visual cues and planning precise placements, such as positioning large objects (\emph{e.g.}, a penguin) to anchor the scene.

    \item \textbf{Tic-Tac-Toe.} The robot perceives the game state, reasons about optimal moves, and places markers to either win or block the opponent. This task evaluates the integration of visual perception, strategic decision-making, and precise placement, requiring real-time dynamic control.
    
    \item \textbf{Make Breakfast Sandwich (Reasoning Control).} The robot assembles a sandwich by coordinating both arms to manipulate bread, ham, and lettuce. It plans the order of ingredients, retrieves items, and completes the sandwich. The task emphasizes long-horizon sequencing, bimanual coordination, and maintaining object stability throughout the process.
    
    \item \textbf{Roast Beef Steak (Reasoning Control).} The robot performs a cooking sequence involving brushing oil, placing the steak in the oven, and operating the oven controls. It coordinates both arms and times actions to ensure proper cooking, evaluating temporal planning and multi-step manipulation.
\end{enumerate}
We carefully constructed reasoning control data for model training. In Visual Rearrangement, the task involves arranging 5 objects across 16 grids, with 800 demonstrations collected for all 80 unique action steps. In Tic-Tac-Toe, the robot plays as red, performing 50 actions per step, resulting in 450 data samples that challenge real-time decision-making and strategic planning. The Make Breakfast Sandwich and Roast Beef Steak tasks rely on control data from section \ref{sec:dexterity}, emphasizing bimanual coordination and long-horizon task execution. All tasks utilize the \textit{interleaved data construction pipeline} from~\cref{fig:statistic}(b)(c), annotating spatial data (bounding boxes, points, and trajectories) to track object positions and scene state, verify task completion, and guide task planning. During training, \ours is fine-tuned with pre-trained weights, combining regression and flow matching.

\noindent\textbf{Reasoning Control with \ours.}
\cref{fig:reasoning_control_example} illustrates qualitative rollouts of \ours{} performing four tasks. In Tic-Tac-Toe, the model perceives the board state, moves to create threats or block the opponent, such as placing a piece at $(0,2)$ to set up circle and later blocking $(1,0)$ to prevent loss. It executes precise piece placements without hesitation. In Visual Rearrangement, \ours{} parses a target layout image, identifies key spatial anchors like placing the gray penguin at $(0,2)$ to define workspace boundaries, and incrementally matches the arrangement while avoiding collisions. In Roast Beef Steak, the model coordinates both arms: brushing oil on the steak, placing it in the oven at the optimal stage (Step~6), closing the door, and pressing the power button. This showcases its temporal reasoning and efficient role allocation between arms. Finally, in Make Breakfast Sandwich, \ours{} plans the assembly by retrieving ingredients—bread, ham, and lettuce—in the correct sequence, finishing with the final bread slice to secure the sandwich fillings.

\begin{figure*}[!ht]
    \centering
    \hspace*{-0.024\textwidth}
    \includegraphics[width=1.04\linewidth]{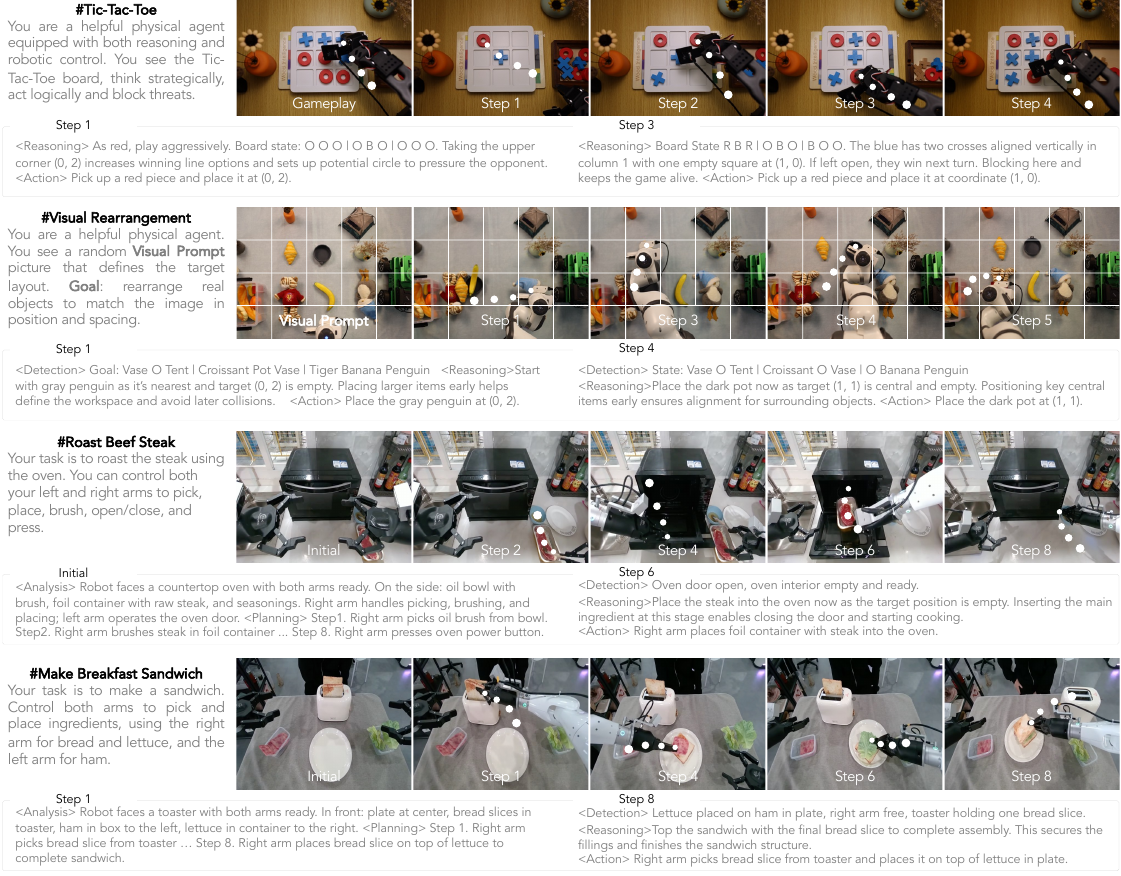}
    \caption{\textbf{Qualitative rollouts for unified reasoning and control.} Each sequence shows how \ours{} perceives the scene, reasons about next steps, and executes precise actions under a single interleaved policy: (a) \textbf{Tic-Tac-Toe}, (b) \textbf{Visual Rearrangement}, (c) \textbf{Roast Beef Steak}, (d) \textbf{Make Breakfast Sandwich}. Planning and execution remain aligned throughout, avoiding plan–act mismatches common in hierarchical pipelines.}
    \label{fig:reasoning_control_example}
\end{figure*}

\noindent\textbf{Quantitative results.}
As shown in~\cref{fig:bar_reasoning_control}, \ours{} achieves the highest average completion score across the four tasks, outperforming GR00T-N1.5 and $\pi_0$. Per-task gains are consistent: Make Breakfast Sandwich (\textbf{84.0\%} vs.\ 70.0\%/$\pi_0$ and 66.0\%/GR00T-N1.5), Roast Beef Steak (\textbf{55.0\%} vs.\ 41.0\% and 46.0\%), Visual Rearrangement (\textbf{79.0\%} vs.\ 66.0\% and 47.0\%), and \textit{Tic-Tac-Toe} (\textbf{76.0\%} vs.\ 24.0\% and 36.0\%). The largest margin occurs in \textit{Tic-Tac-Toe} (+40.0 points over the best baseline), reflecting superior game-state reasoning and real-time execution. The next largest is in Make Breakfast Sandwich (+14.0 points), showing strong temporal sequencing and precise control.

We attribute these gains to the alignment of \textbf{ natural characteristics} achieved by the interleaved vision-text-action training, which unifies symbolic planning and continuous control within a cohesive backbone. This eliminates the interface gap between planner and controller, reducing error propagation and enabling smooth, context-aware action execution. Qualitative sequences confirm that \ours{} maintains coherent task strategies while performing fine-grained and stable manipulations, which is an essential capability for robust agents embodied in the open world.

\begin{figure*}[htbp]
    \centering
    \hspace*{-0.05\textwidth}
    \includegraphics[width=1.0\linewidth]{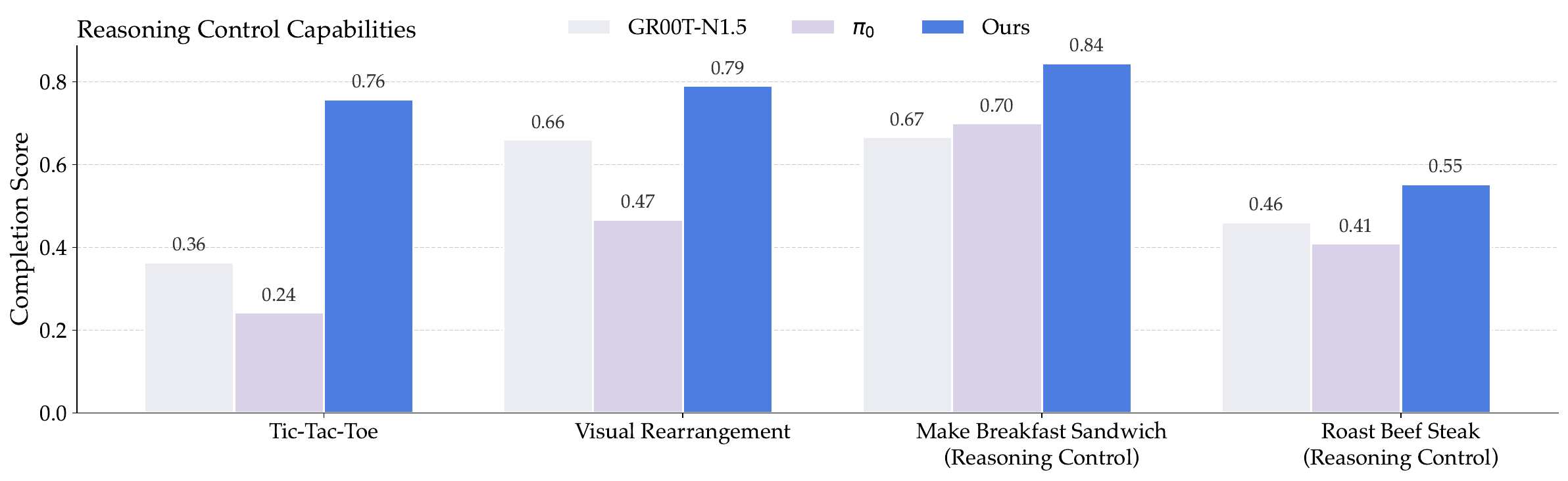}
    \caption{\textbf{Success rates on the Reasoning–control benchmark.} \ours{} outperforms both hierarchical baselines on all tasks and on average, with the largest gains in \textit{Tic-Tac-Toe} and \textit{Roast Beef Steak}.}
    \label{fig:bar_reasoning_control}
\end{figure*}

\subsection{Discussion: Architecture and Data Contributions to Generalization}
In this section, we examine how the unified architecture and interleaved multimodal pre-training affect model generalization along the three axes defined in \cref{sec:generalization}: \emph{visual}, \emph{action}, and \emph{language}. Our objective is to isolate the contributions of (i) a hybrid decoding mechanism that combines discrete autoregression with continuous flow matching under shared parameters, and (ii) large-scale, interleaved vision–text–action supervision, to overall robustness and scaling behavior in the open world. The \textbf{benchmark settings} include:
\begin{enumerate}[leftmargin=*, topsep=0pt, noitemsep]
    \item \textbf{LIBERO}~\citep{liu2023libero}: Evaluates the model’s capacity to fit training distributions and execute long-horizon, multi-stage manipulation, stressing policy learning and temporal credit assignment. 
    \item \textbf{Agibot Sandwich}, a single real-world bimanual task requiring precise sequencing of bread, ham, and lettuce placement—used to assess the impact of interleaved embodied data on single-task.
    \item \textbf{WidowX Generalization} with BridgeData V2~\citep{walke2023bridgedata}: Five task groups selected from the WidowX 250 S \textbf{Prepare Vegetables} and \textbf{Arrange Cups} suites. We adopt the generalization protocols in \cref{sec:generalization}: \emph{Visual} (changes in object appearance, background, lighting), \emph{Action} (altered object positions, novel viewpoints, new instances), and \emph{Language} (typos, rephrasings, fuzzy descriptions). For training, we randomly sample relevant demonstrations from BridgeData at four scales: 50, 500, 5k, and 50k.
\end{enumerate}
For the \textbf{model configurations}, we explore the following settings:
\begin{enumerate}[leftmargin=*, topsep=0pt, noitemsep]
    \item \ours (fast): A fully autoregressive baseline using the Fast-tokenizer~\citep{pertsch2025fast}, trained on Bridge Data with interleaved vision–text–action pairs.
    \item \ours (base): Our proposed unified hybrid decoding model that is trained solely on robot control data. discrete autoregressive decoding for vision language tokens and continuous flow-matching denoising for robotic actions, sharing parameters across modalities.
    \item \ours (llava): \ours co-trained on robot control data with general vision–language instruction datasets (LLaVA-Instruct-150K, RefCOCO) to improve grounding and natural language understanding.
    \item \ours (interleaved): \ours trained with interleaved vision–text–action sequences from corresponding data subsets, emphasizing \emph{Physical Common Sense}, \emph{Spatial Understanding}, \emph{State Estimation}, and \emph{Task Reasoning}.
\end{enumerate}

\begin{figure*}[htbp]
    \centering
    \hspace*{-0.025\textwidth}
    \includegraphics[width=1.02\linewidth]{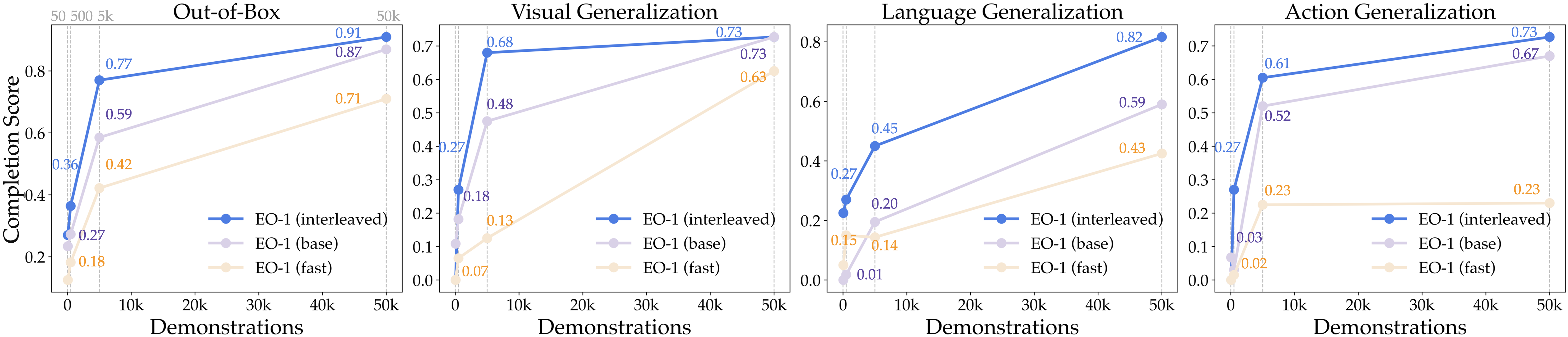}
    \caption{WidowX generalization performance across data scales for \ours (base), \ours (interleaved), and \ours (fast) with interleaved vision–text–action data.}
    \label{fig:ablation_demonstration}
\end{figure*}

\begin{table*}[!h]
    \centering
    \resizebox{1.0\textwidth}{!}{
        \begin{tabular}{lccccc}
            \toprule
            \bf LIBERO Benchmark (Arch)      & \bf LIBERO-Spatial  & \bf LIBERO-Object         & \bf LIBERO-Goal             & \bf LIBERO-Long           & \bf Overall \\
            \midrule
            \ours (fast)                     & 93.3\%              & 95.5\%                    & 88.0\%                      & 74.0\%                    & 88.0\%      \\
            \rowhighlight
            \ours (base)                           & \bf 99.7\%          & 99.8\%                    & 99.2\%                      & 94.8\%                    & 98.2 \%     \\
            \midrule
            \bf Agibot Sandwich (Data)       & \bf In-distribution & \bf Visual Generalization & \bf Language Generalization & \bf Action Generalization & \bf Overall \\
            \midrule
            \ours (base)                            & 0.85                & 0.83                      & 0.66                        & 0.75                      & 0.77        \\
            \ours (llava)                    & 0.84                & 0.44                      & 0.62                        & 0.73                      & 0.66        \\
            \ours (interleaved)                      & 0.89                & 0.88                      & 0.83                        & 0.75                      & 0.84        \\
            \midrule
            \bf WidowX Generalization (Data) & \bf Out-of-Box      & \bf Visual Generalization & \bf Language Generalization & \bf Action Generalization & \bf Overall \\
            \midrule
            \ours (base)                           & 0.87                & 0.73                      & 0.59                        & 0.67                      & 0.71        \\
            \ours (llava)                    & 0.37                & 0.37                      & 0.35                        & 0.28                      & 0.34        \\
            \rowhighlight
            \ours (interleaved)                      & 0.91                & 0.73                      & 0.82                        & 0.73                      & 0.80        \\
            \bottomrule
        \end{tabular}
    }
    \caption{LIBERO, WidowX and Agibot evaluation for \ours (base), \ours (interleaved), and \ours (fast) with interleaved vision–text–action data.}
    \label{tab:ablation_mmu_fast}
\end{table*}

\noindent\textbf{Unified Decoding Across Modalities Outperforms Autoregression.}
As shown in~\cref{fig:ablation_demonstration}, our hybrid model (\ours (base)) consistently outperforms fully autoregressive \ours (fast) across all generalization regimes and data scales. With only 50 demonstrations, \ours achieves up to +0.25 absolute gains in \emph{Action Generalization}, and the margin persists at scale, reaching 0.91 Out-of-Box at 50k demos while pure autoregression plateaus. In particular, pure AR does not show emerging \emph{Language Generalization}, stalling at 0.43 accuracy even as data grow, and its \emph{Action Generalization} saturates at 0.23 from 5 to 50k demonstrations. Similar patterns appear in~\cref{tab:ablation_mmu_fast}: in LIBERO, \ours (fast) achieves 88.0\% overall success rate compared to 98.2\% for \ours (base), underscoring the precision gains from incorporating flow-matching into action decoding. Across tasks, discrete-only decoding struggles to capture low-frequency action modes and fails to generalize robustly under distribution shift, even with large-scale data. In contrast, our hybrid architecture unifies discrete autoregression with continuous flow matching, enabling more accurate control and scaling-driven improvements that emerge strongly in open-world scenarios.

\noindent\textbf{Scaling Interleaved Vision–Text–Action Data Boosts Generalization.}
Scaling interleaved demonstrations from 50 to 50k (~\cref{fig:ablation_demonstration}) shows that \ours (interleaved) consistently outperforms \ours (base), with faster and more robust generalization as the data grows. This is especially evident in \emph{Language Generalization}, where \ours (interleaved) improves significantly from 0.27 to \textbf{0.82}, compared to \ours (base), which increases from 0.01 to \textbf{0.59}. This gap highlights the advantage of incorporating multimodal data aligned with both language and control tasks, enabling faster learning and more effective generalization. However, not all multimodal data are beneficial. As shown in~\cref{tab:ablation_mmu_fast}, incorporating generic instruction-following data, such as \ours (llava), leads to a performance drop, from \textbf{0.77} to \textbf{0.66} on Agibot Sandwich and from \textbf{0.71} to \textbf{0.34} on WidowX Generalization. This suggests that \textit{multimodal data misaligned with embodied control tasks biases the model toward linguistic priors, weakening its physical grounding}. We observe that \textbf{\textit{Out of the Action Domain}} generalization struggles on limited datasets, despite dense multimodal annotations. In Agibot Sandwich, performance stays at 0.75 for both the base model and \ours (interleaved). In contrast, WidowX, trained on large-scale, diverse Bridge Data, improves significantly (0.73 vs. 0.67), highlighting how multimodal data in multi-task settings enhances action generalization. These results underscore the need for finely annotated, task-aligned multimodal datasets incorporating physical knowledge for effective embodied control.

Finally, The superior performance of our unified architecture, which combines discrete autoregression for vision–language understanding with continuous flow matching for action generation, demonstrates how joint modeling enhances both control precision and generalization. This approach not only strengthens temporal alignment between multimodal understanding and motor control but also ensures stable, high-performance outcomes even in open-world, distribution-shifted scenarios, proving the power of integrating multiple modalities under a unified framework.
\section{Related Work}
\label{sec:related_work}

\textbf{Vision-Language-Action Models.}
Vision-language-action models (VLAs) have emerged as a promising approach for generalizable robot control by extending pre-trained vision-language models (VLMs) to action prediction. Autoregressive approaches~\citep{brohan2023rt,kim2024openvla,qu2025spatialvla,pertsch2025fast,liu2024rdt} discretize continuous actions into tokens with linear bins, spatial grids or quantized DCT byte-pair encoding, naturally aligning with VLM training paradigms and exhibiting inherent advantages in knowledge transfer from web-scale pre-training. Despite these advantages, this discrete paradigm suffers from inference latency and limited resolution for high-frequency control tasks. Alternative works~\citep{team2024octo,liu2024rdt,black2024pi_0,bjorck2025gr00t} introduce continuous action experts via diffusion or flow-matching~\citep{ho2020denoising,lipman2022flow,song2025hume}, achieving fast inference and dexterous robotic control. Several efforts~\citep{liu2025hybridvla,kim2025fine} attempt to combine discrete autoregression and continuous denoising, employing collaborative action ensemble or two-stage tuning procedure. However, these models suffer from the challenges of knowledge preservation and training efficiency, as the gradients from the action expert degrade the pre-trained VLM backbone~\citep{driess2025knowledge,intelligence2025pi_0_5}.

Co-training strategies have emerged as a promising solution that integrates diverse data sources and enables generalization to new environments. $\pi_{0.5}$~\citep{intelligence2025pi_0_5} employs a two-stage co-training approach: first pre-training in multimodal data sources with FAST tokenized actions, then tuning the action expert for low-level control. Building on this foundation, ~\citep{driess2025knowledge} formalizes a single-stage recipe insulating the VLM backbone during VLA training. Complementary works~\citep{lin2025onetwovla,zhou2025chatvla} further explore reasoning and generalization through $\pi_0$-style co-training or purely autoregressive paradigms. Nevertheless, instruction following capabilities remain suboptimal, fundamentally due to architectural bottlenecks and the fact that current approaches are predominantly trained on separate image-text pairs or robotic episodes data, overlooking the rich temporal dynamics and causal relationships inherent in embodied interactions. Addressing these challenges necessitates better co-training strategies, improved cross-modal transfer mechanisms, and larger-scale interleaved datasets that comprehensively capture the full spectrum of robotic episodes.

\noindent\textbf{Unified Multimodal Models.} Unifying understanding and generation~\citep{deng2025emerging,chen2025janus,liao2025mogao} has witnessed remarkable progress and emergent properties beyond individual specialized models. Pioneering unified multimodal models~\citep{zhou2024transfusion,team2024chameleon,xie2024show} integrate these capabilities through autoregressive or diffusion modeling in a single architecture, fusing both text autoregression and visual generation in the shared backbone. Recent advancements~\citep{deng2025emerging,ma2025unitok,chen2025janus,xie2025show} introduce modality-specific paramaters or dual visual encoding to optimize training pipelines, facilitate joint token semantic and superior performance. Alternative studies~\citep{chen2025blip3,dong2023dreamllm,ge2024seed,pan2025transfer} explored unified models with frozen LLMs, assemble LMMs backbones and visual generative models via adapters or learnable queries. While assembling frozen LLMs struggles to demonstrate promising capabilities in robotic learning~\citep{kim2024openvla,bjorck2025gr00t,qu2025spatialvla}, unifying understanding and generation inspires architecture design in representative works~\citep{black2024pi_0,intelligence2025pi_0_5,bjorck2025gr00t} and exhibits significant emergent open-world capabilities.
\section{Conclusion and Future Work}
\label{sec:conclusion}
In this article, we present \ours, a fully open training recipe to foster the research community to develop an advanced embodied foundation model. \ours is composed of \ours model, \dataset dataset, and \benchmark benchmark, where \dataset is collected by a scalable pipeline for curating interleaved vision-text-action data to improve \ours model's open-world generalization.
\ours effectively synergizes auto-regressive decoding and flow matching denoising in a single unified model that is co-trained with web data and curated interleaved embodied data \dataset, enabling seamless perception, planning, reasoning, and acting in complex physical environments.
Our evaluations across embodied reasoning and robot control experiments demonstrate that \ours consistently outperforms strong vision–language–action baselines, exhibiting superior multimodal reasoning and dexterous robot control capabilities in open-world generalization.
Moreover, \benchmark presents a new benchmark to evaluate embodied reasoning capabilities of embodied foundation models.
In summary, \ours is fully open-sourced to the community and achieves a forward step for equipping general-purpose autonomous robots with human-like reasoning and acting abilities.

\ours has made a solid step toward developing general embodied AI, featuring with seamless embodied reasoning and action in the open world.
While the results with \ours demonstrate promising robot control capabilities, there remains key aspects that future work can address.
First, we aim to enhance \ours's reasoning and action ability to handle complex scenarios involving navigation, obstacle avoidance, failure detection and analysis, human intent recognition, and human-robot interaction/cooperation. This requires seamlessly integrate more general multimodal understanding, embodied reasoning, and action abilities into one system, leading to easier-to-use robots in real life. Second, we plan to explore a more efficient design of a unified model architecture or an asynchronous inference pipeline to enable human-like simultaneous reasoning and action. Finally, we will incorporate more data and robot embodiments (\emph{e.g.}, human data and humanoid robots) into the training recipe to enable \ours to be a broader foundation for general-purpose autonomous robots.

\bibliography{main}

\clearpage
\appendix
\section{Contributors and Acknowledgments}
\label{sec:contributors}
\subsection{Core Contributors}
Delin Qu, Haoming Song, Qizhi Chen, Zhaoqing Chen, Xianqiang Gao, Dong Wang, Bin Zhao

\noindent{\textbf{{Model Architecture and Training}} Delin Qu, Haoming Song, Qizhi Chen

\noindent{\textbf{Real-Robot Experiments}} Haoming Song, Delin Qu, Haoran Yang, Jiacheng Bao, Qizhi Chen

\noindent{\textbf{Simulation and Benchmarking}} Delin Qu, Haoming Song, Xianqiang Gao, Xinyi Ye, Yiwen Tang

\noindent{\textbf{Data and Benchmark Curation}} Zhaoqing Chen, Xianqiang Gao, Qizhi Chen, Delin Qu, Junli Liu

\noindent{\textbf{Research Leads}} Dong Wang, Bin Zhao

\subsection{Contributors}
Qi Lv, Xinyi Ye, Modi Shi, Guanghui Ren, Cheng Ruan, Maoqing Yao, Haoran Yang, Jiacheng Bao.

\subsection{Acknowledgments}
This work is supported by Shanghai Artificial Intelligence Laboratory. We acknowledge the AgiBot team, including Modi Shi, Cheng Ruan, Guanghui Ren, and Maoqing Yao, for providing hardware support and maintenance of the AgiBot G-1 robots. Special thanks to Xinyi Ye for video editing contributions and Qizhi Chen for webpage design.
\\
\\
\section{Training Dataset Statistics}
\label{sec:training_dataset_statistics}
\begin{table}[htbp]
    \vspace{-2ex}
    \footnotesize
    \centering
    \resizebox{1.0\textwidth}{!}{
    \begin{tabular}{lllllll}
        \toprule
        Dataset                & Samples & Tokens  & Duration (hr) & FPS & Camera View                  & Category                  \\
        \midrule
        LLaVA-Video-178K       & 2.6M    & 5.95B   & --            & --  & Third-person                 & Web multimodal data       \\
        LLaVA-1.5              & 1.1M    & 390.45M & --            & --  & Third-person                 & Web multimodal data       \\
        Pixmo-Points           & 1.8M    & 538.92M & --            & --  & Third-person                 & Web multimodal data       \\
        RoboVQA                & 0.2M    & 218.29M & --            & --  & Egocentric                   & Web multimodal data       \\
        RefCOCO                & 50.1K   & 9.46M   & --            & --  & Third-person                 & Web multimodal data       \\
        \midrule
        Agibot-Beta            & 213.8M  & 116.17B & 2560.7        & 30  & Egocentric, left, right      & Real robot                \\
        DROID (OXE)            & 23.1M   & 8.41B   & 428.3         & 15  & Left, Right, wrist           & Real robot                \\
        RT-1 (OXE)             & 3.7M    & 579.32M & 338.4         & 3   & Egocentric                   & Real robot                \\
        Bridge-v2 (OXE)        & 2.0M    & 1.89M   & 111.1         & 5   & Shoulder, left, right, wrist & Real robot                \\
        Robo-Mind              & 2.1M    & 2.09M   & 19.3          & 30  & Egocentric, left, right      & Real robot                \\
        SO100-Community        & 7.9M    & 1.19B   & 72.1          & 30  & Third-person, wrist          & Real robot                \\
        IPEC-Franka            & 0.8M    & 108.52M & 19.7          & 10  & Third-person, wrist          & Real robot                \\
        \midrule
        \dataset               & 1.5M    & 1.0B    & 13.9          & 30  & Egocentric, left, right      & Interleaved Embodied Data \\
        \midrule
        Total multimodal data  & 6.2M    & 7.1B    & --            & --  & --                           & --                        \\
        Total robot data       & 253.4M  & 127.3B  & --            & --  & --                           & --                        \\
        Total Interleaved data & 1.5M    & 1.0B    & 13.9          & 30  & Egocentric, left, right      & Interleaved Embodied Data \\
        \midrule
        Total                  & 260.6M  & 135.4B  & --            & --  & --                           & --                        \\
        \bottomrule
        \end{tabular}
    }
    \caption{\centering Pre-training Dataset Statistics.}
    \vspace{-3ex}
    \label{tab:pretraining_dataset_stats}
\end{table}
\clearpage
\section{\ours Reasoning Examples}
\begin{figure*}[!ht]
  \centering
  \vspace{-3ex}
  \includegraphics[width= 1\linewidth]{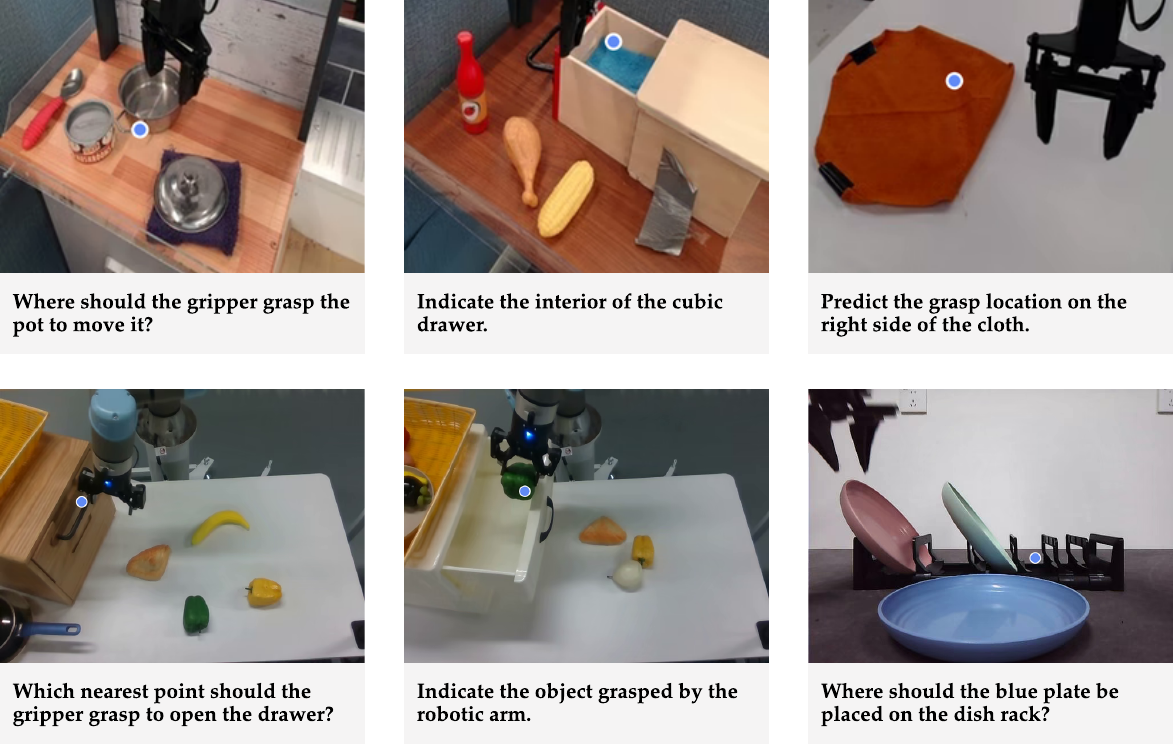}
  \caption{\textbf{Object pointing examples.}}
  \vspace{-3ex}
  \label{fig:reasoning_examples_points}
\end{figure*}
\begin{figure*}[!ht]
  \centering
  \includegraphics[width= 1\linewidth]{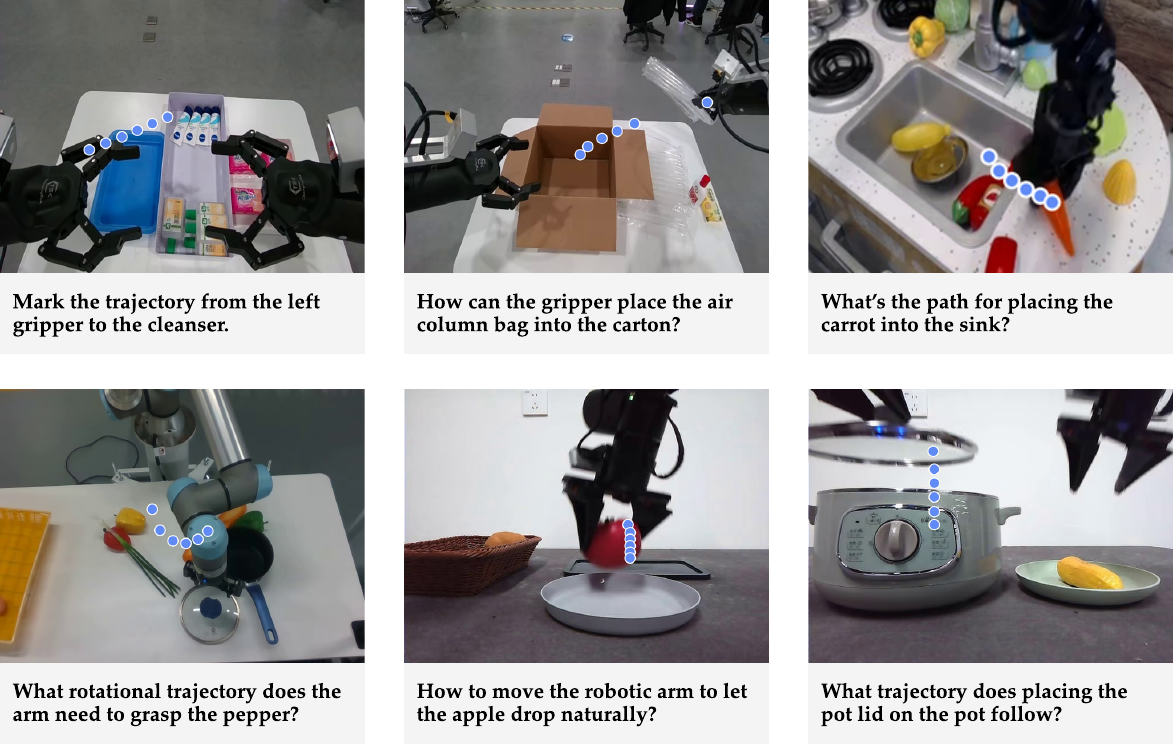}
  \caption{\textbf{Trajectory prediction examples.}}
  \label{fig:reasoning_examples_traj}
\end{figure*}

\begin{figure*}[htbp]
  \centering
  \includegraphics[width= 1\linewidth]{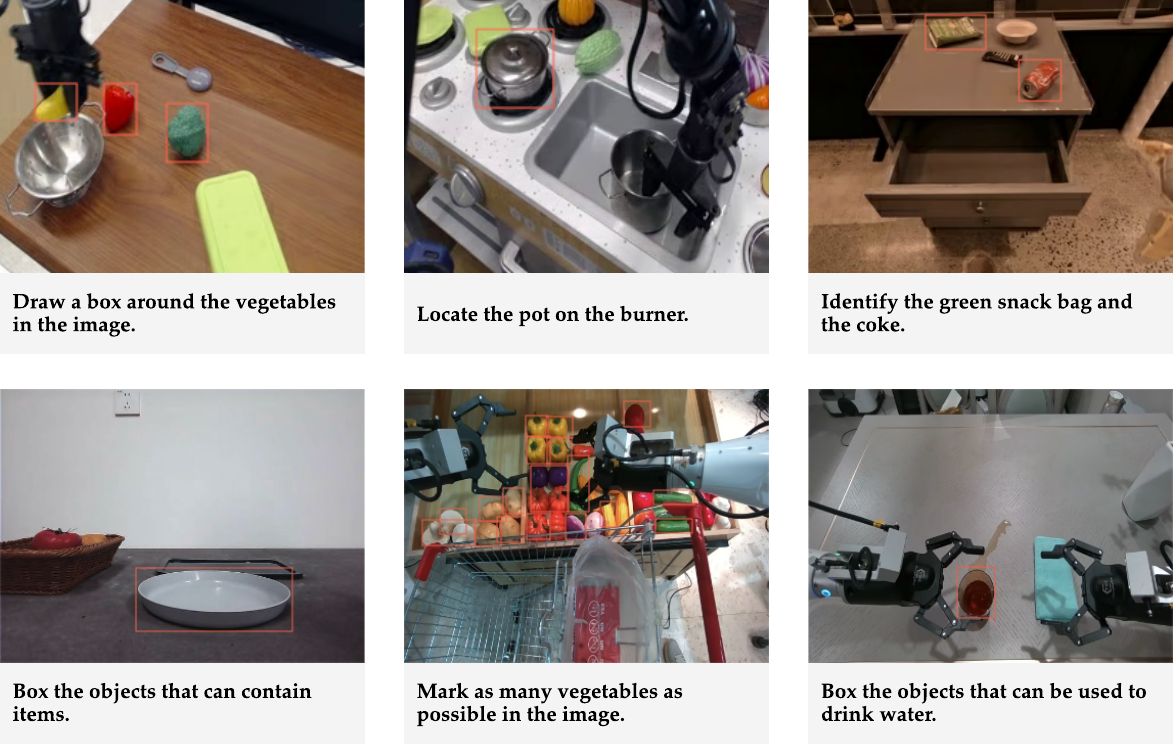}
  \caption{\textbf{Object referring examples.}}
  \label{fig:reasoning_examples_ref}
\end{figure*}

\begin{figure*}[htbp]
  \centering
  \includegraphics[width= 1\linewidth]{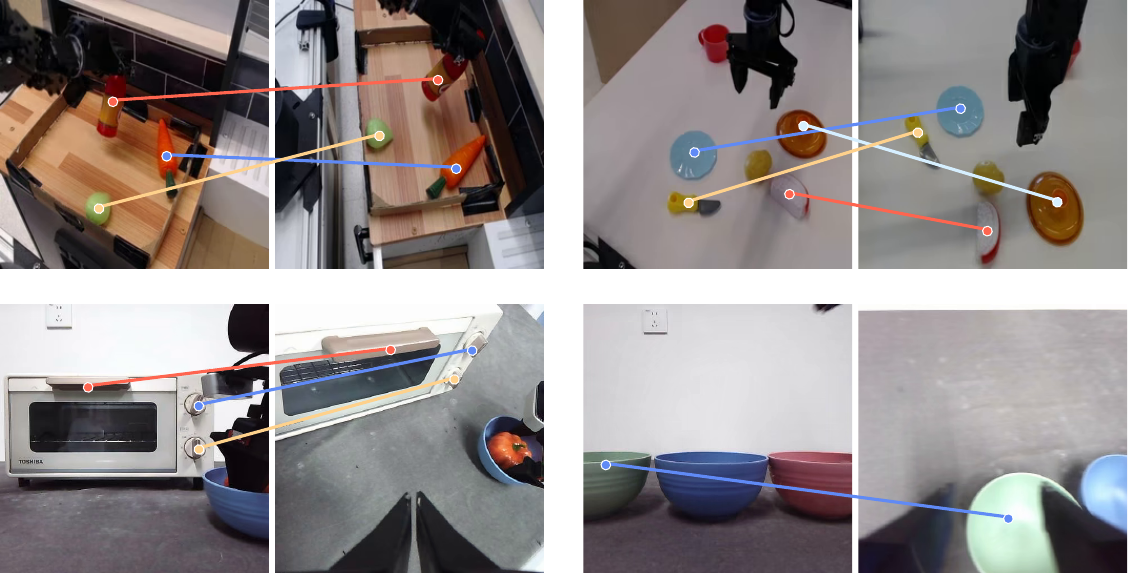}
  \caption{\textbf{Multiview reasoning examples.}}
  \label{fig:reasoning_examples_mtviw}
\end{figure*}

\begin{figure*}[htbp]
  \centering
  \includegraphics[width= 1\linewidth]{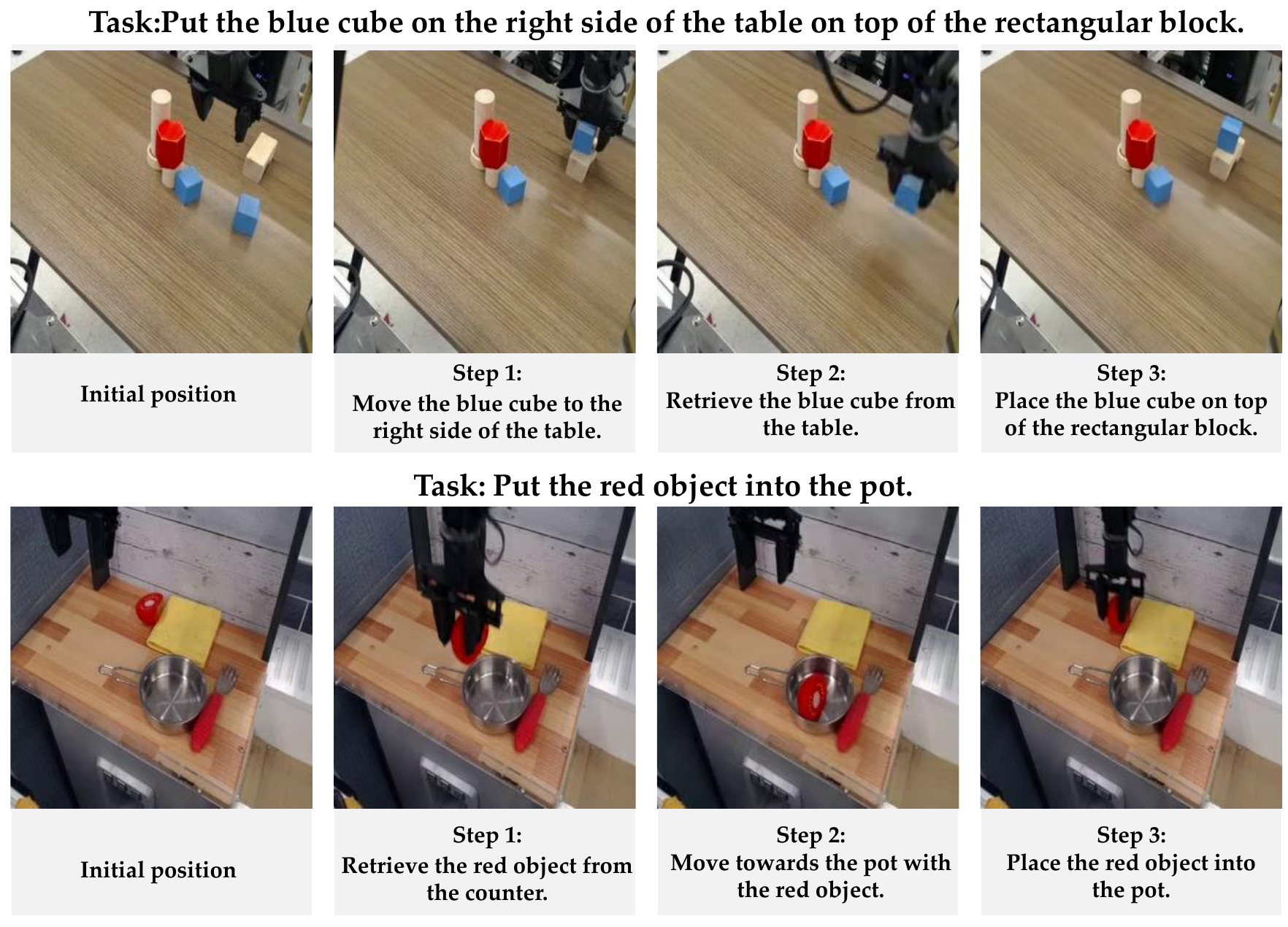}
  \caption{\textbf{Task planning examples.}}
  \label{fig:reasoning_examples_planning}
\end{figure*}

\begin{figure*}[htbp]
  \centering
  \includegraphics[width= 1\linewidth]{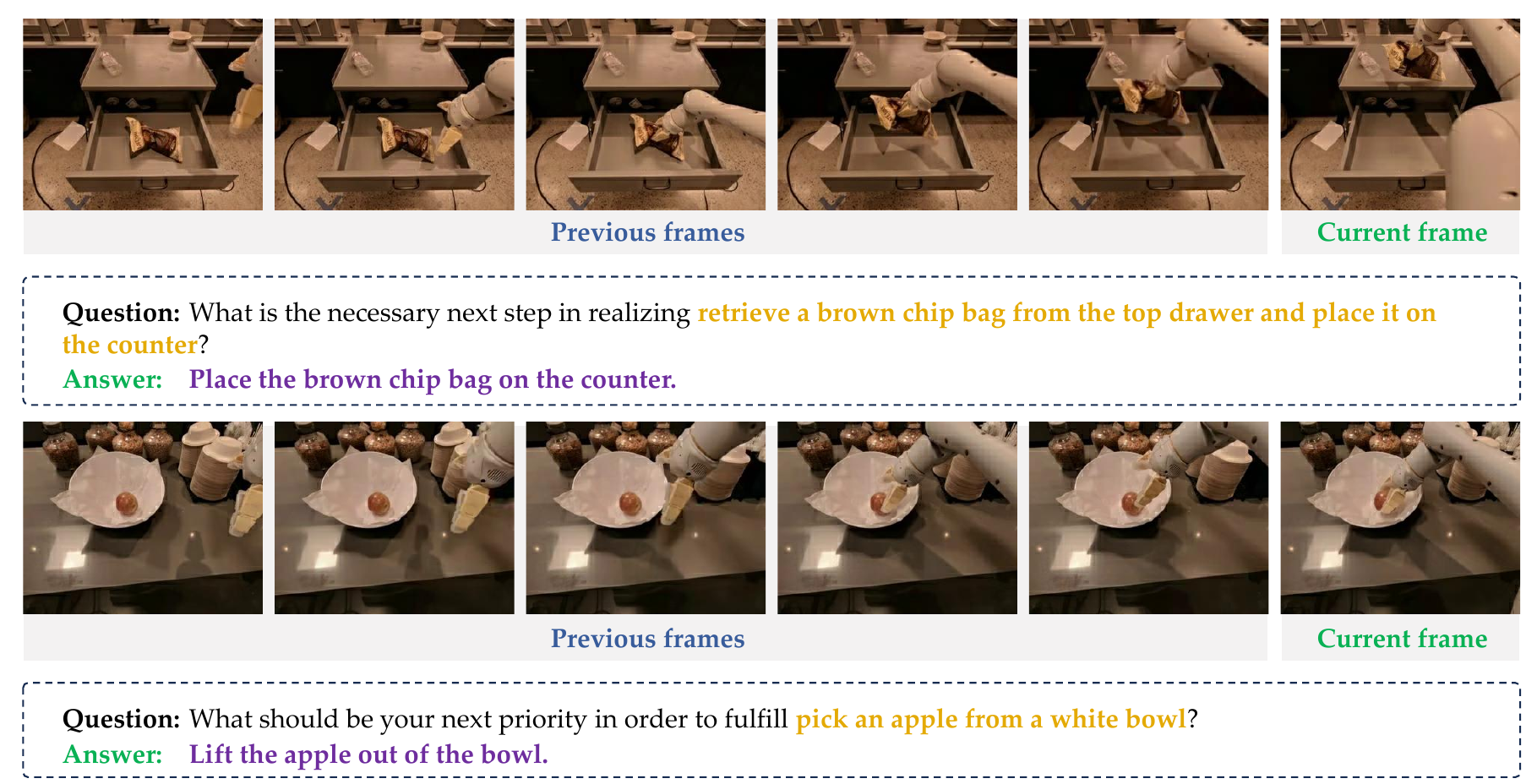}
  \caption{\textbf{Subtask QA examples.}}
  \label{fig:reasoning_examples_subtask_qa}
\end{figure*}

\begin{figure*}[htbp]
  \centering
  \includegraphics[width= 0.95\linewidth]{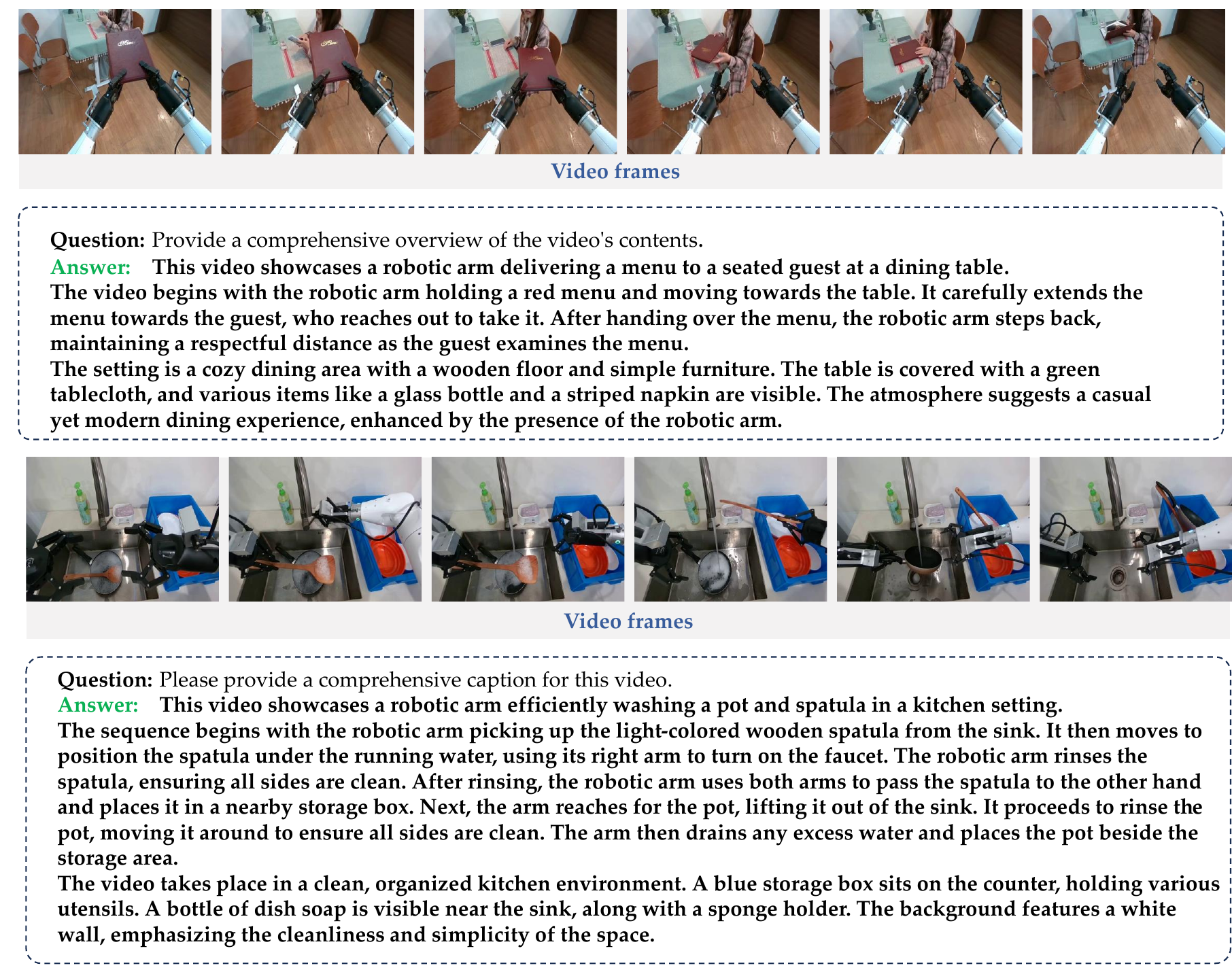}
  \caption{\textbf{Episode caption examples.}}
  \label{fig:reasoning_examples_episode_caption}
\end{figure*}

\begin{figure*}[htbp]
  \centering
  \includegraphics[width= 0.95\linewidth]{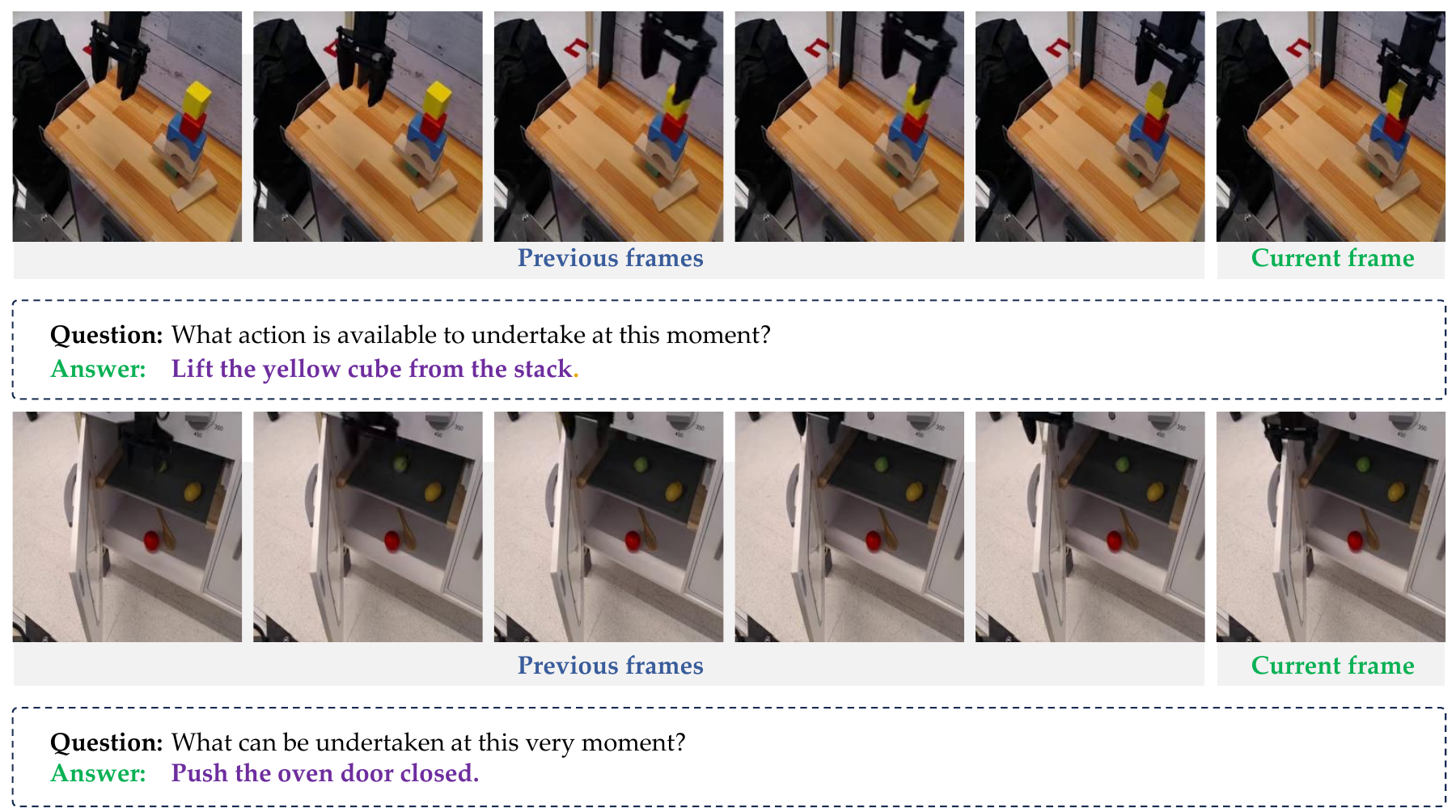}
  \caption{\textbf{Affordance QA examples.}}
  \label{fig:reasoning_examples_affordance_qa}
\end{figure*}

\begin{figure*}[!ht]
  \centering
  \includegraphics[width= 0.95\linewidth]{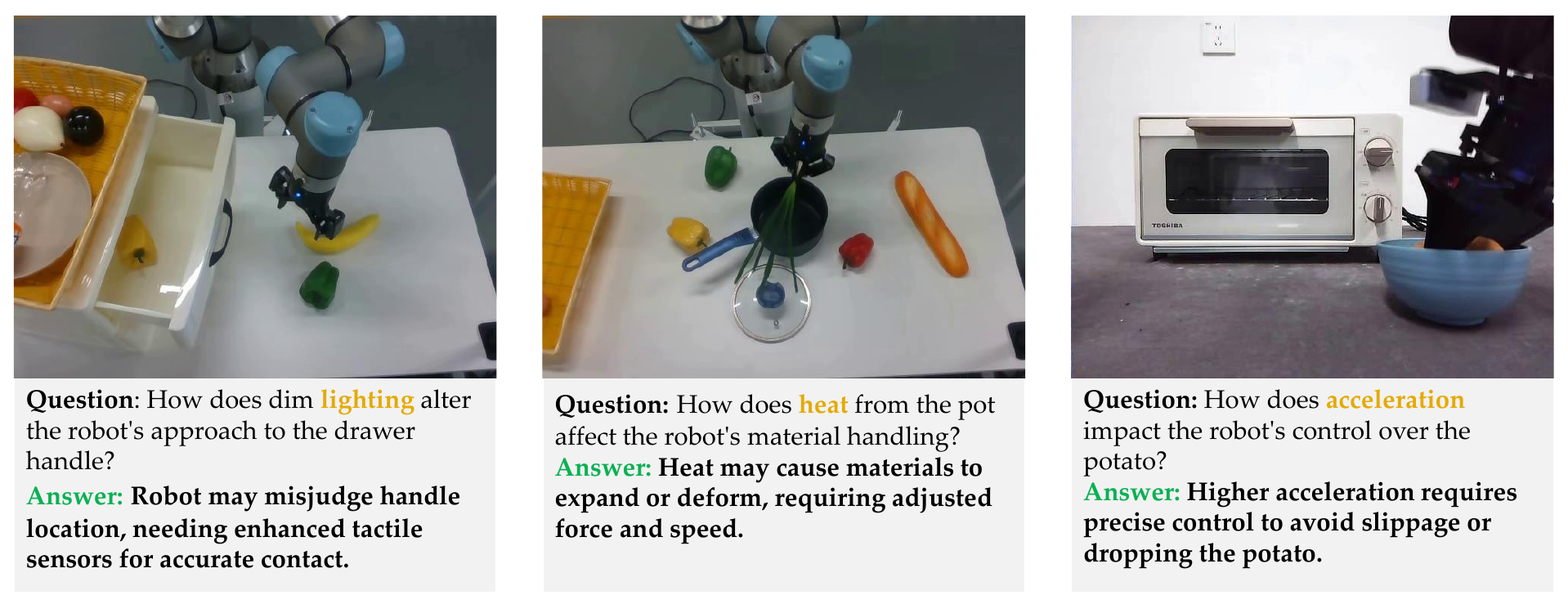}
  \caption{\textbf{Physical common sense examples.}}
  \label{fig:reasoning_examples_physical_common_sense}
\end{figure*}

\begin{figure*}[!ht]
  \centering
  \includegraphics[width= 0.95\linewidth]{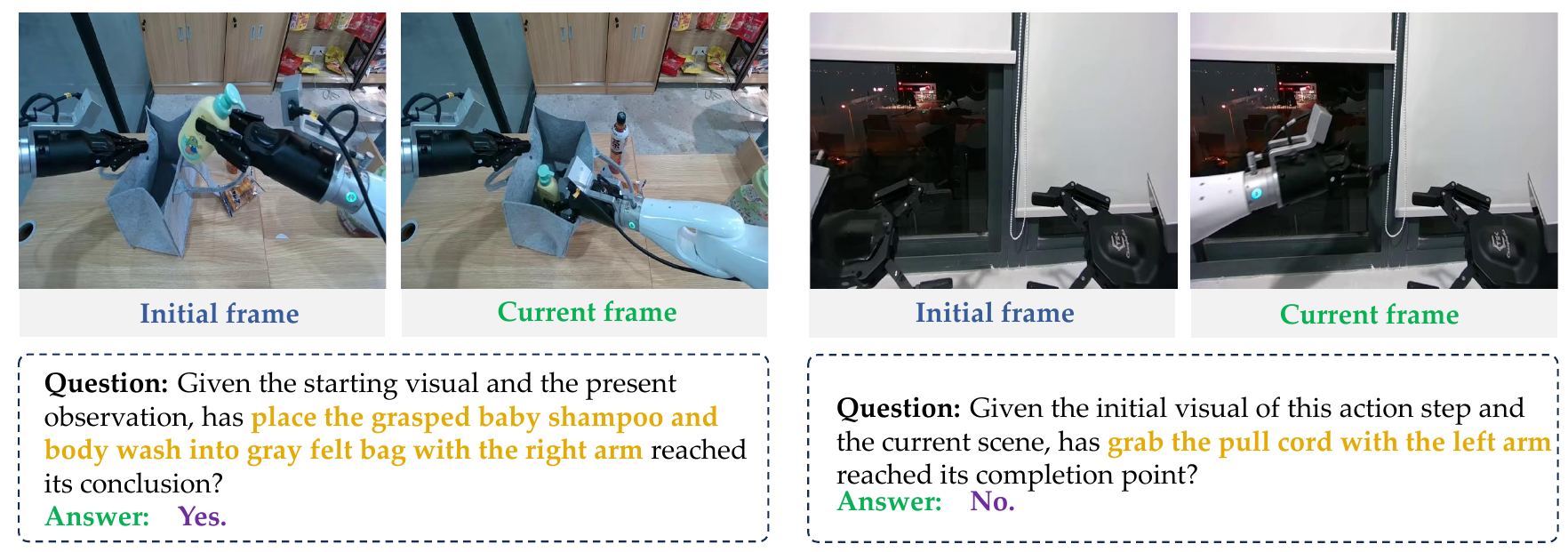}
  \caption{\textbf{Failure detection examples.}}
  \label{fig:reasoning_examples_failure_detection}
\end{figure*}

\begin{figure*}[h]
  \centering
  \includegraphics[width= 0.95\linewidth]{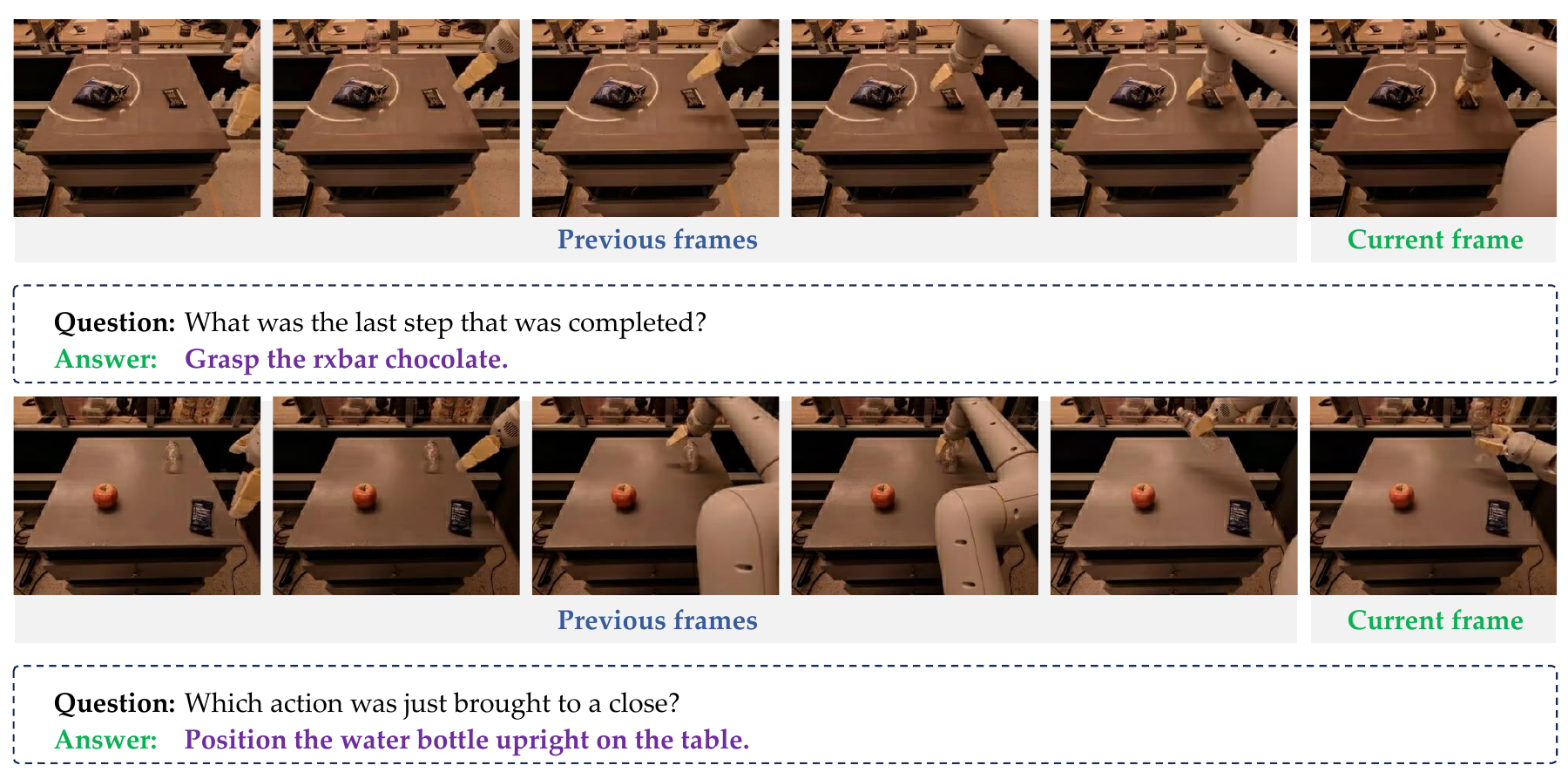}
  \caption{\textbf{Process verification examples.}}
  \label{fig:reasoning_examples_process_verification}
\end{figure*}

\clearpage
\section{Data Annotation Pipeline Details}
\label{sec:data_preprocessing}
In this section, we provide additional details of the data preprocessing pipeline, which complements the description in the main paper. The process is divided into four major components: i) video filtering and curation, ii) video splitting and captioning, iii) QA generation for temporal and spatial reasoning, and iv) cleaning and rewriting.
\subsection{Video Filtering and Curation}
The first step focuses on ensuring the diversity and quality of robot videos. Since existing robot datasets often consist of highly redundant scenes collected in constrained environments, we employ a feature-based clustering strategy to enhance visual variety. Specifically, pretrained visual backbones are used to extract video features, followed by K-means clustering. From each cluster, we select a balanced set of videos, and further apply human-assisted inspection to remove samples from the same scene. \cref{fig:web_screenshot_cluster} illustrates the interface for this process, where clusters are visualized and manually pruned to improve dataset coverage.

\begin{figure*}[!ht]
  \centering
  \includegraphics[width= 1\linewidth]{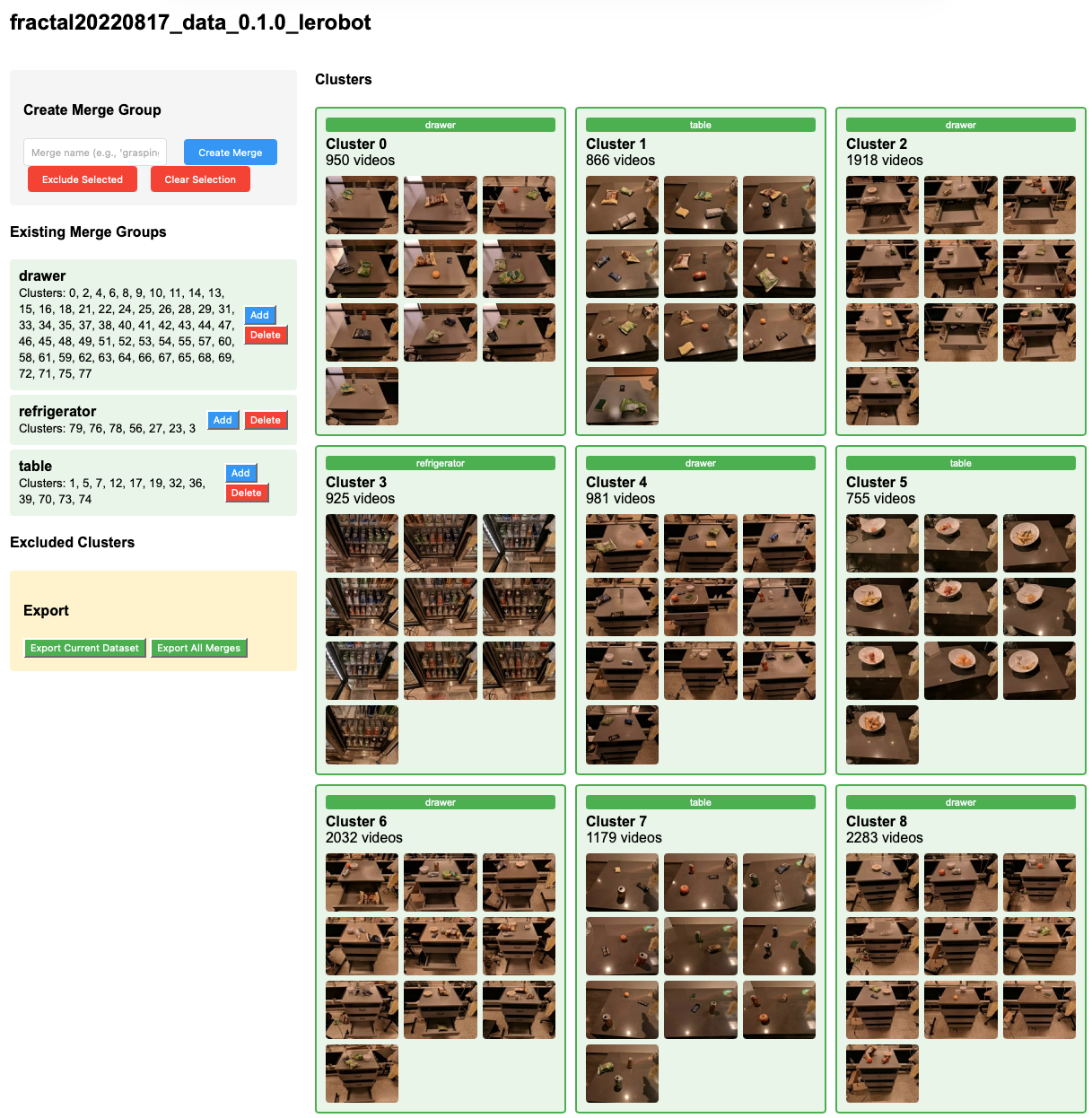}
  \caption{\textbf{Web interface for video filtering and clustering.} We extract features using a pretrained backbone and apply K-means clustering to group visually similar videos. Human verification is then used to remove redundant samples and ensure diversity.}
  \label{fig:web_screenshot_cluster}
\end{figure*}

To further enrich the data for tasks requiring multiview or cross-scene reasoning, we organize all available observations into a unified interface (\cref{fig:web_screenshot_mulvw_selection}). This allows selecting videos with sufficient viewpoint diversity, ensuring that multiview reasoning tasks are well-supported.

\begin{figure*}[!ht]
  \vspace{-4ex}
  \centering
  \includegraphics[width= 1\linewidth]{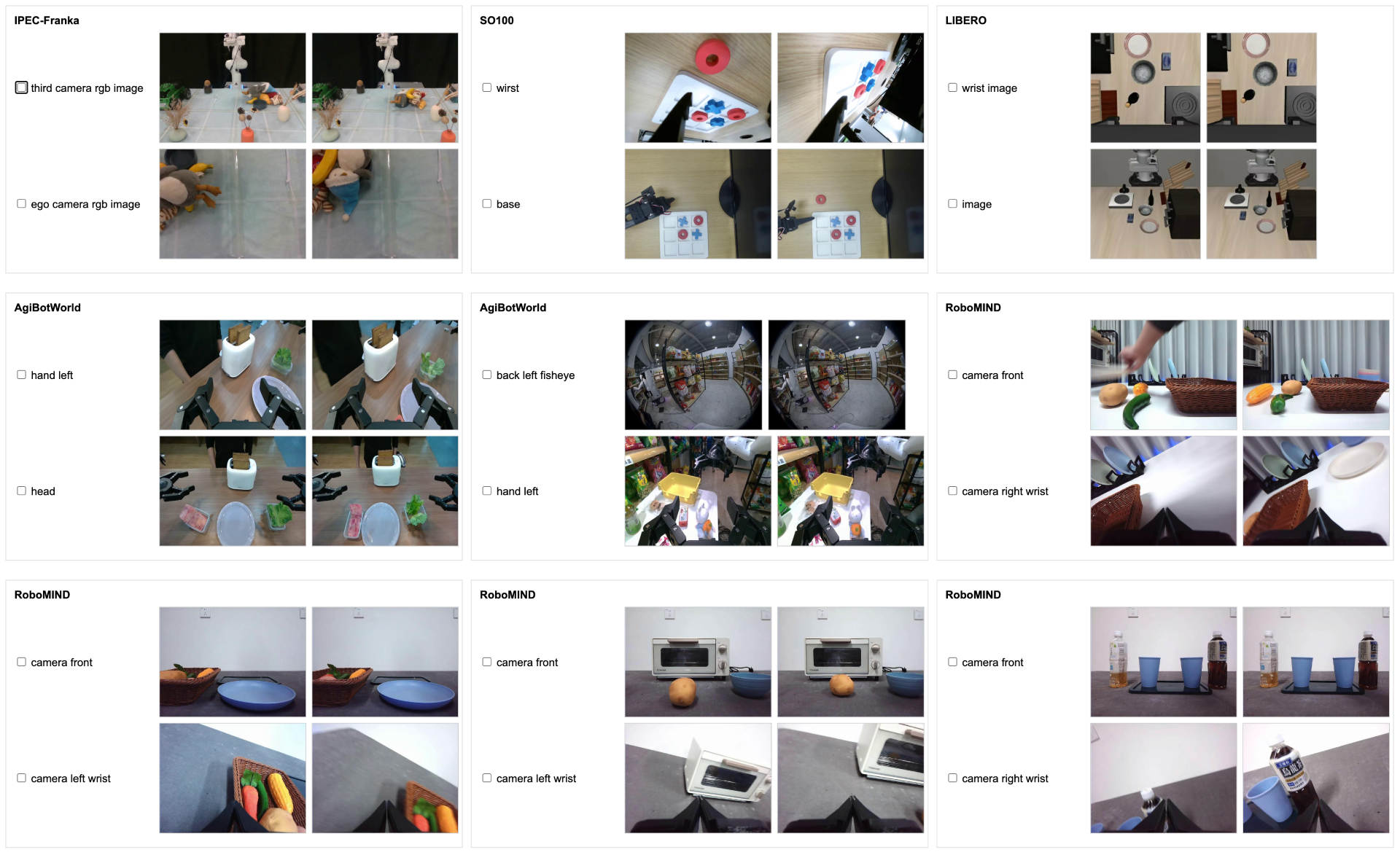}
  \caption{\textbf{Manual selection of multiview videos.} All observations from different datasets are displayed, and the interface allows selecting samples with sufficient viewpoint diversity for multiview reasoning tasks.}
  \label{fig:web_screenshot_mulvw_selection}
\end{figure*}

\subsection{Video Splitting and Captioning}
Once a diverse video set is curated, we process them into finer-grained clips. Annotators or pretrained VLMs are instructed to segment videos into meaningful short clips, each containing a single subtask. For each clip, descriptive captions are generated to capture the specific robot actions. These captions serve dual purposes: they support video captioning QA data and also provide contextual prompts for subsequent reasoning annotations. \cref{fig:video_caption_prompt} and \ref{fig:clip_caption_prompt} show examples of the prompts used to guide caption generation.
\begin{figure*}[htbp]
  \centering
  \vspace{-4ex}
  \includegraphics[width= 1\linewidth]{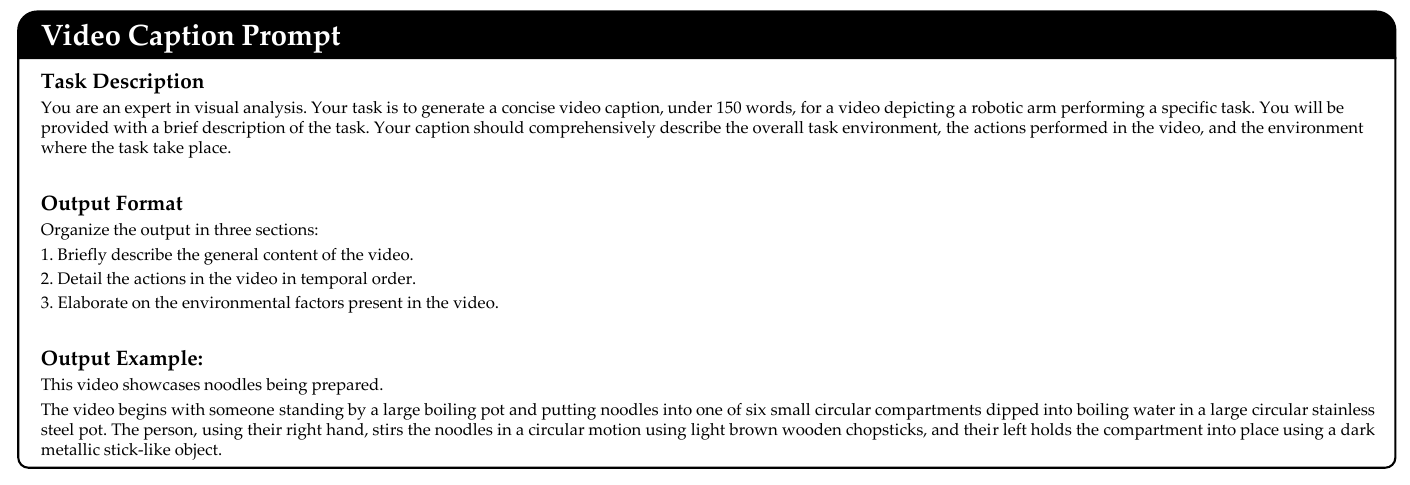}
  \caption{\textbf{Video caption prompt}.}
  \vspace{-4ex}
  \label{fig:video_caption_prompt}
\end{figure*}
\begin{figure*}[ht]
  \vspace{-2ex}
  \centering
  \includegraphics[width= 1\linewidth]{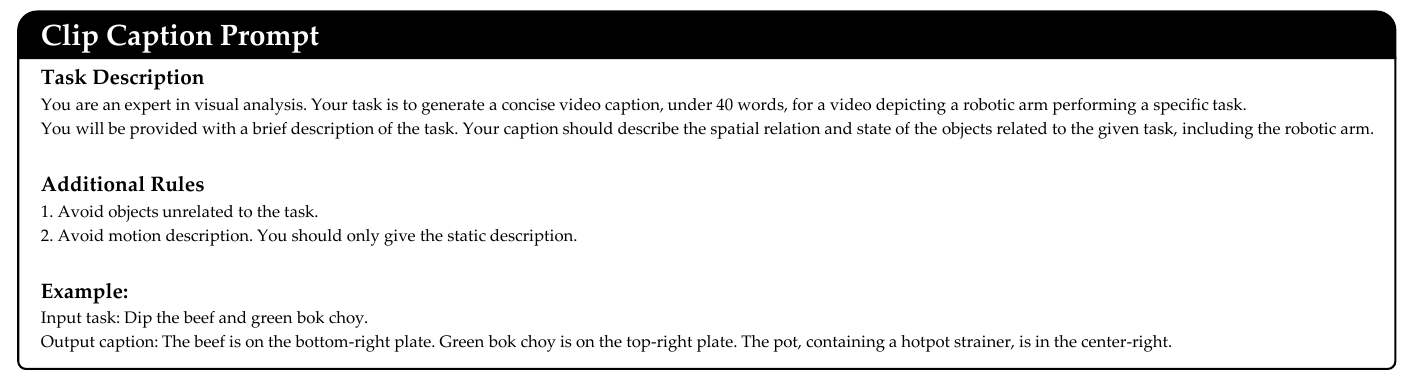}
  \caption{\textbf{Clip caption prompt}.}
  \label{fig:clip_caption_prompt}
\end{figure*}

\subsection{QA Generation for Temporal and Spatial Reasoning}
The next stage is to construct question-answer pairs that capture both temporal and spatial reasoning abilities. We adopt a prompting strategy, where VLMs are guided with carefully designed templates to generate diverse questions, and human annotators refine the final answers.

\subsubsection{Temporal Reasoning}
\label{sec:temporal_prompts}
For temporal reasoning, we focus on two key aspects: \emph{task planning} and \emph{physical common sense}. Task planning questions require understanding the sequential structure of subtasks, while physical common sense questions evaluate the model’s ability to reason about object dynamics and feasibility in the physical world. \cref{fig:task_planning_prompt} and \ref{fig:physical_question_prompt} provide examples of the prompts used to elicit these two categories of questions.

\begin{figure*}[htbp]
    \centering
    \includegraphics[width= 1\linewidth]{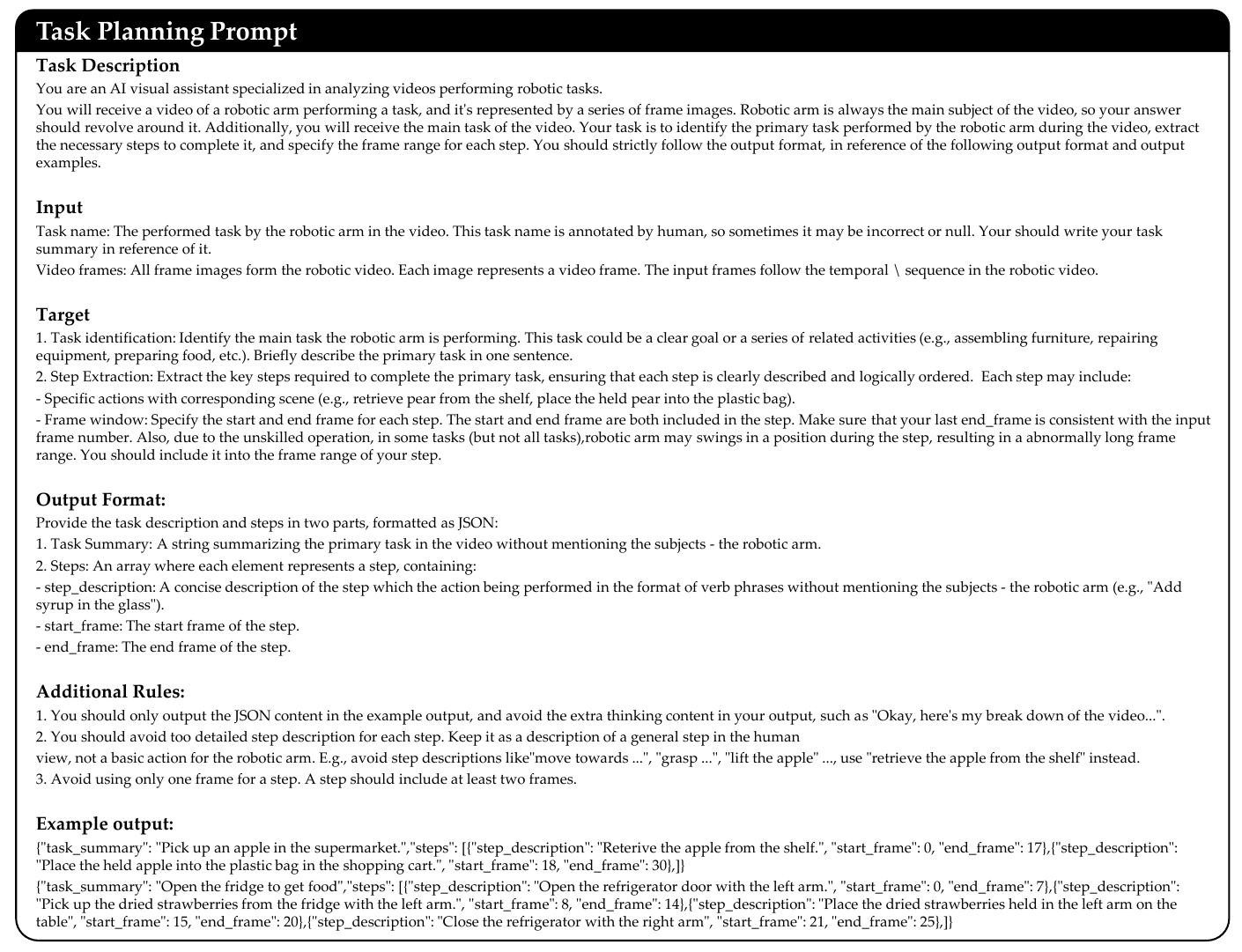}
    \caption{\textbf{Task planning prompt}.}
    \label{fig:task_planning_prompt}
\end{figure*}

\begin{figure*}[htbp]
    \centering
    \includegraphics[width= 1\linewidth]{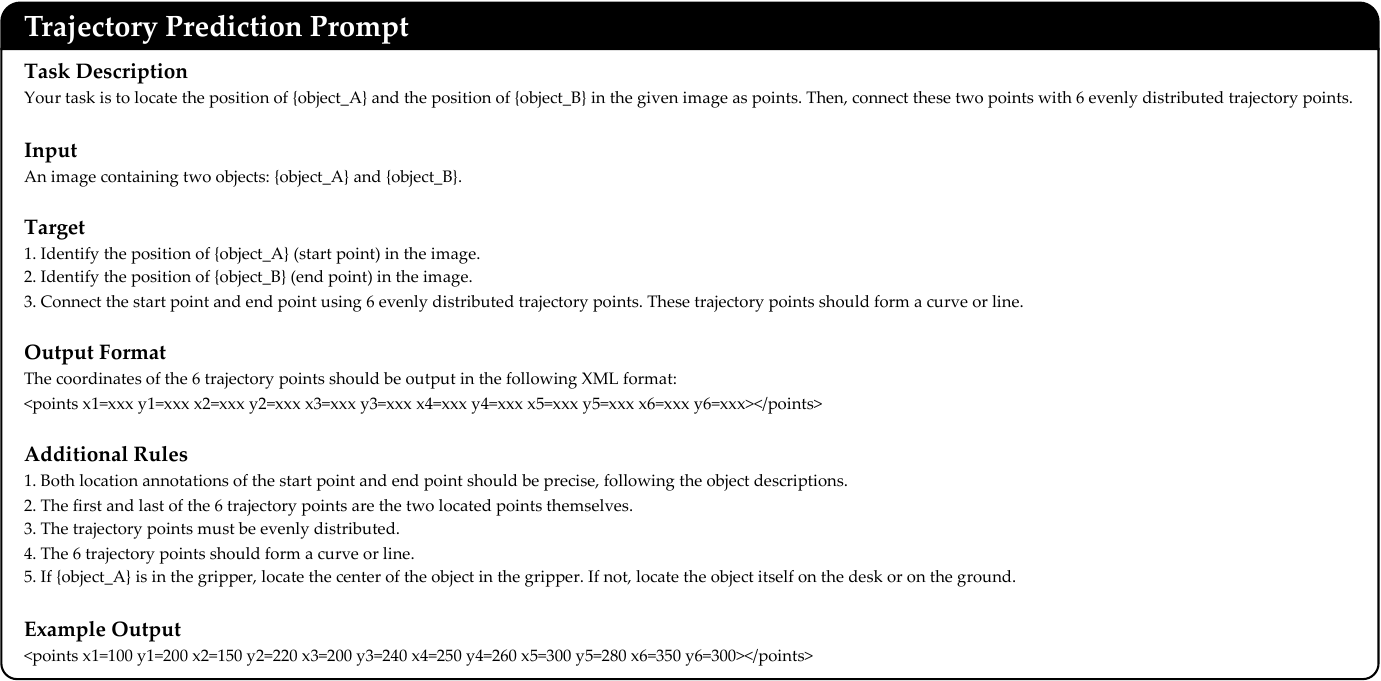}
    \caption{\textbf{Trajectory prediction prompt}.}
    \label{fig:trajectory_prediction_prompt}
\end{figure*}

\begin{figure*}[t]
    \vspace{-2ex}
    \centering
    \includegraphics[width= 1\linewidth]{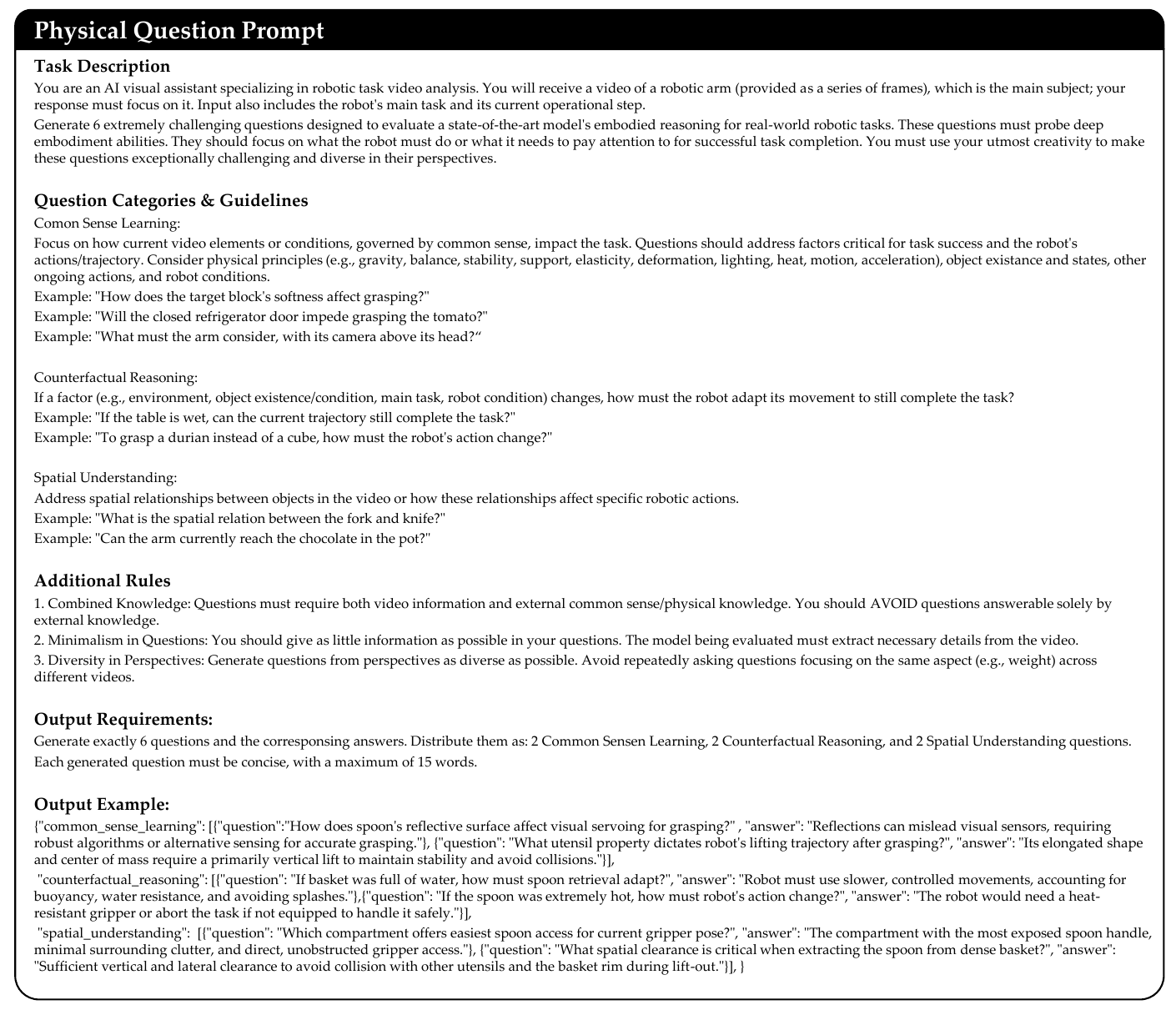}
    \caption{\textbf{Physical question prompt}.}
    \label{fig:physical_question_prompt}
    \vspace{-2ex}
\end{figure*}

\subsubsection{Spatial Reasoning}
\label{sec:spatial_prompts}
For spatial reasoning, we construct four sub-tasks: trajectory prediction, object pointing, object referring, and multiview reasoning. The multiview data are generated using the same prompting strategy as object pointing, without a separate prompt design. \cref{fig:trajectory_prediction_prompt}, \ref{fig:object_pointing_prompt}, and \ref{fig:object_referring_prompt} show representative prompts used to construct these spatial reasoning tasks. Together, they ensure that the dataset captures fine-grained spatial relations necessary for embodied intelligence.

\begin{figure*}[h]
    \vspace{-2ex}
    \centering
    \includegraphics[width= 1\linewidth]{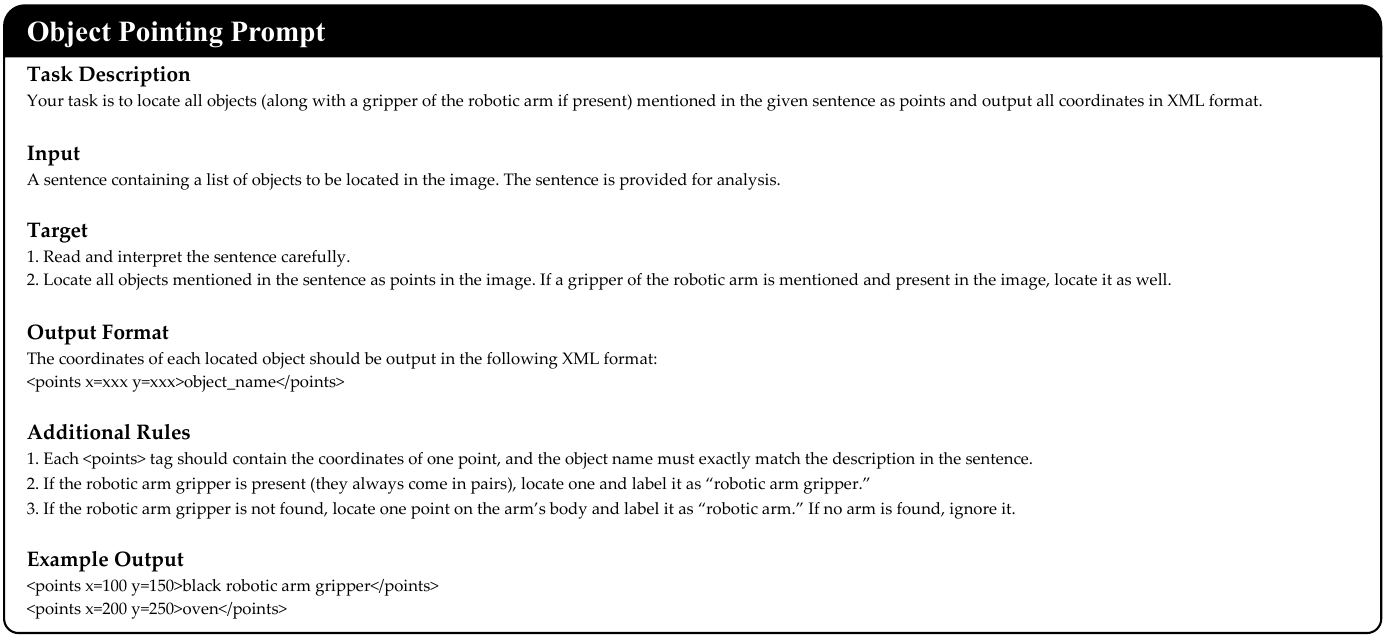}
    \caption{\textbf{Object pointing prompt}.}
    \label{fig:object_pointing_prompt}
    \vspace{-2ex}
\end{figure*}

\begin{figure*}[htbp]
    \vspace{-2ex}
    \centering
    \includegraphics[width= 1\linewidth]{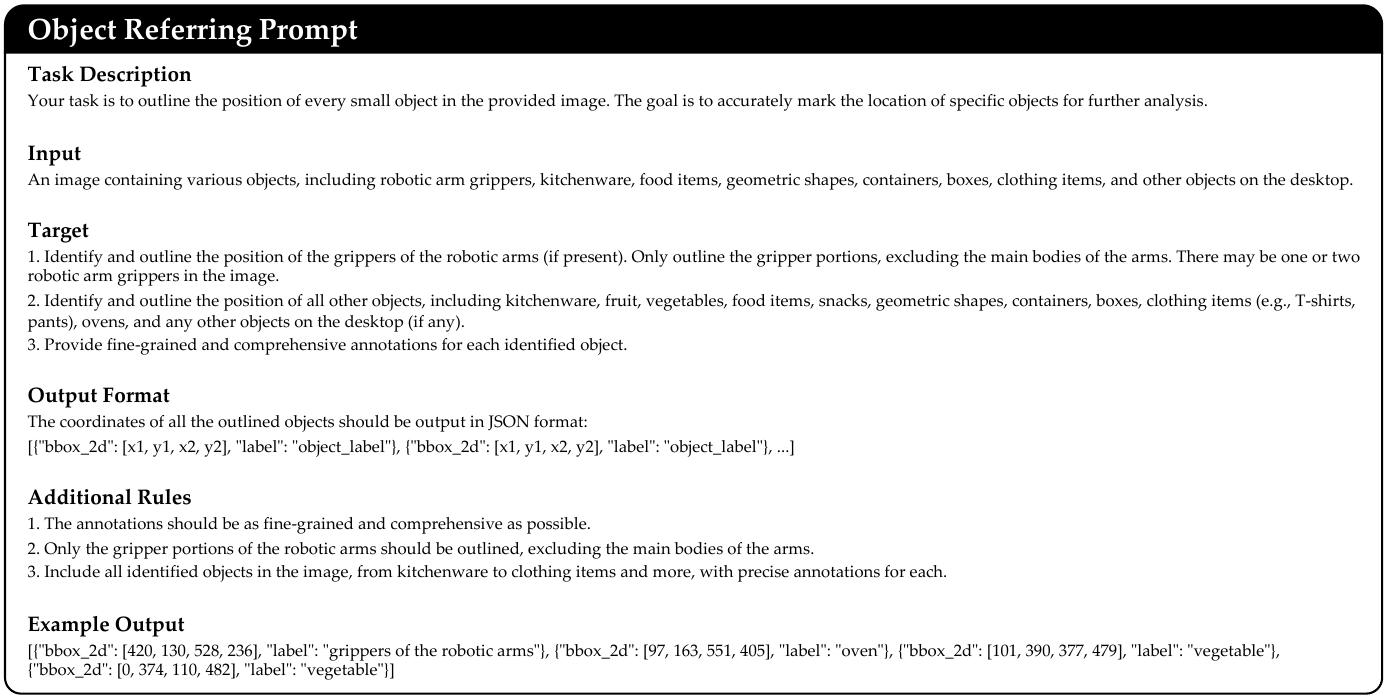}
    \caption{\textbf{Object referring prompt}.}
    \label{fig:object_referring_prompt}
    \vspace{-2ex}
\end{figure*}

\paragraph{Cleaning and Rewriting}
Finally, to enhance linguistic quality and ensure answer validity, we perform a post-processing step. Rule-based filtering is applied to remove invalid or ambiguous spatial answers, and an LLM is prompted to rewrite QA pairs. This step increases textual diversity and introduces natural fuzziness, making the dataset more robust. An example of the rewriting prompt is shown in \cref{fig:semantic_deblurring_prompt}.

\begin{figure*}[h]
    \centering
    \includegraphics[width= 1\linewidth]{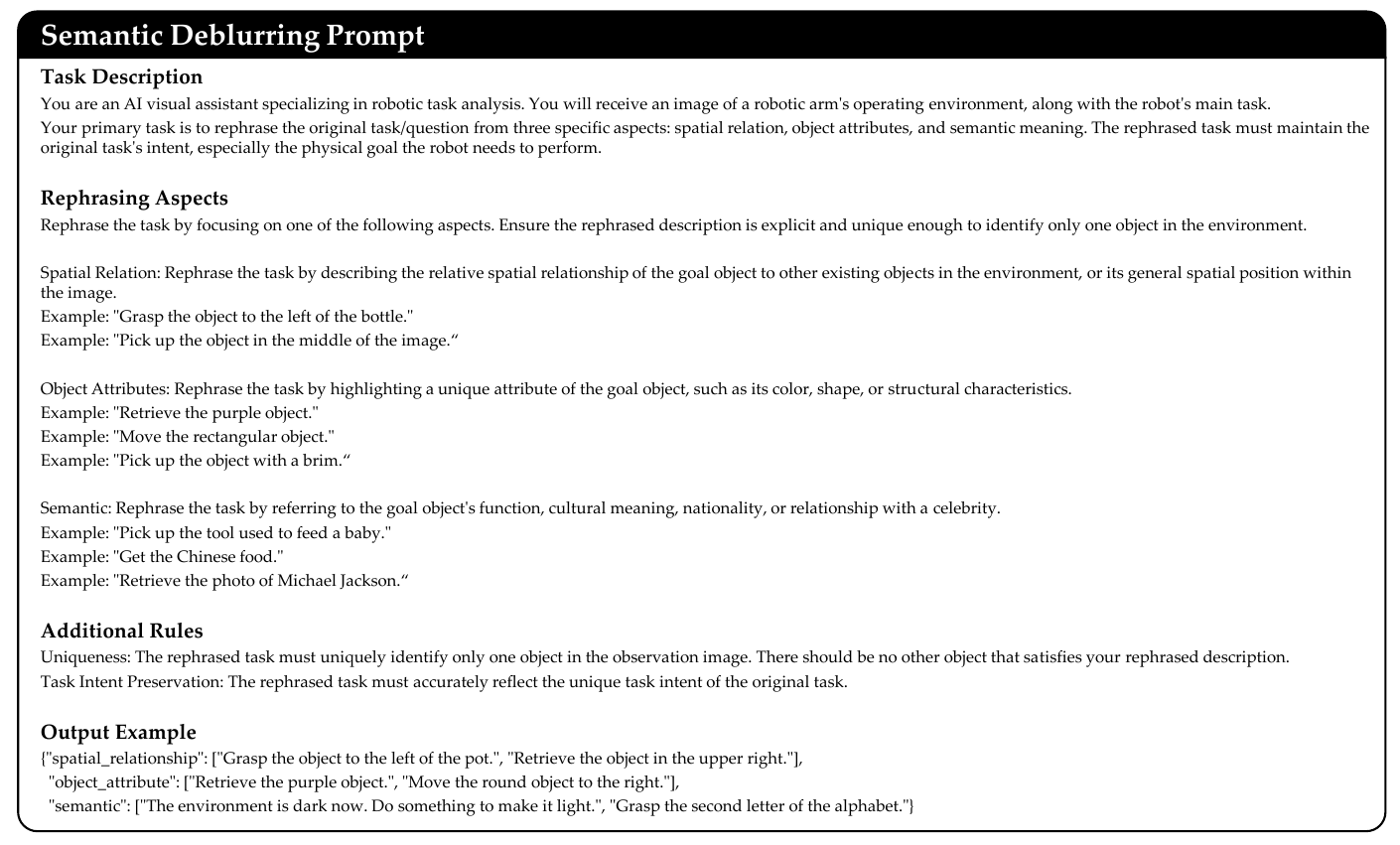}
    \caption{\textbf{Semantic deblurring prompt}.}
    \label{fig:semantic_deblurring_prompt}
\end{figure*}

\end{document}